\documentclass[runningheads]{llncs}

\usepackage{eccv}

\usepackage{lipsum} 
\usepackage{eccvabbrv}

\usepackage{graphicx}
\usepackage{booktabs}

\usepackage[accsupp]{axessibility}  

\usepackage[utf8]{inputenc}
\usepackage{textcomp}
\usepackage{newunicodechar}
\newunicodechar{ρ}{\ensuremath{\rho}}
\newunicodechar{ω}{\ensuremath{\omega}}
\newunicodechar{γ}{\ensuremath{\gamma}}
\newunicodechar{≠}{\ensuremath{\neq}}
\newunicodechar{∞}{\ensuremath{\infty}}
\newunicodechar{μ}{\textmu}

\usepackage{hyperref}

\usepackage[hypcap=true]{caption}

\usepackage{orcidlink}
\usepackage{tabularx}
\newcolumntype{Y}{>{\centering\arraybackslash}X}

\usepackage{xcolor}

\usepackage{amssymb}

\usepackage{multirow} 

\usepackage{pifont, xcolor}
\newcommand{\cmark}{\textcolor[rgb]{0.35,0.50,0.10}{\ding{51}}}
\newcommand{\xmark}{\textcolor[rgb]{0.49,0.47,0.62}{\ding{55}}}

\usepackage{float}
\usepackage{fvextra}
\usepackage{mdframed}
\fvset{
  fontsize=\footnotesize,
  breaklines=true,
  breakanywhere=true,
  numbers=none,
  xleftmargin=0pt
}

\definecolor{sagegrey}{HTML}{A7A787}
\definecolor{dustypurple}{HTML}{79769E}
\definecolor{darkpurple}{HTML}{5C59A0}
\definecolor{ochre}{HTML}{D9DC8F}

\definecolor{lightpurple}{HTML}{EEEDFE}   
\definecolor{lightsage}{HTML}{F4F2EA}     
\definecolor{lightochre}{HTML}{F5F2CC}    

\newcolumntype{Z}[1]{>{\hsize=#1\hsize\centering\arraybackslash}X}

\begin{document}

\title{OCR-Based Field Extraction for Archaeological Pottery Metadata: The CENTURIA Dataset} 

\titlerunning{CENTURIA: Transcription and Field Extraction for Pottery Records}

\author{Gissu Valentina Naghavi \inst{1}\orcidlink{0009-0002-8512-7635} \and
Dominik Hagmann\inst{2}\orcidlink{0000-0002-4481-6234} \and
Martin Kampel\inst{1}\orcidlink{0000-0002-5217-2854} \and
Irene Ballester\inst{1}\orcidlink{0000-0002-0219-9063}}

\authorrunning{G.V.~Naghavi et al.}


\institute{Computer Vision Lab, TU Wien, Vienna, Austria \\
\email{\{gissu.naghavi,martin.kampel,irene.ballester\}@tuwien.ac.at} \and
Austrian Archaeological Institute, Austrian Academy of Sciences, Vienna, Austria\\
\email{\{dominik.hagmann\}@oeaw.ac.at}}

\maketitle

\begin{center}
    \centering

    \includegraphics[width=0.93\linewidth]{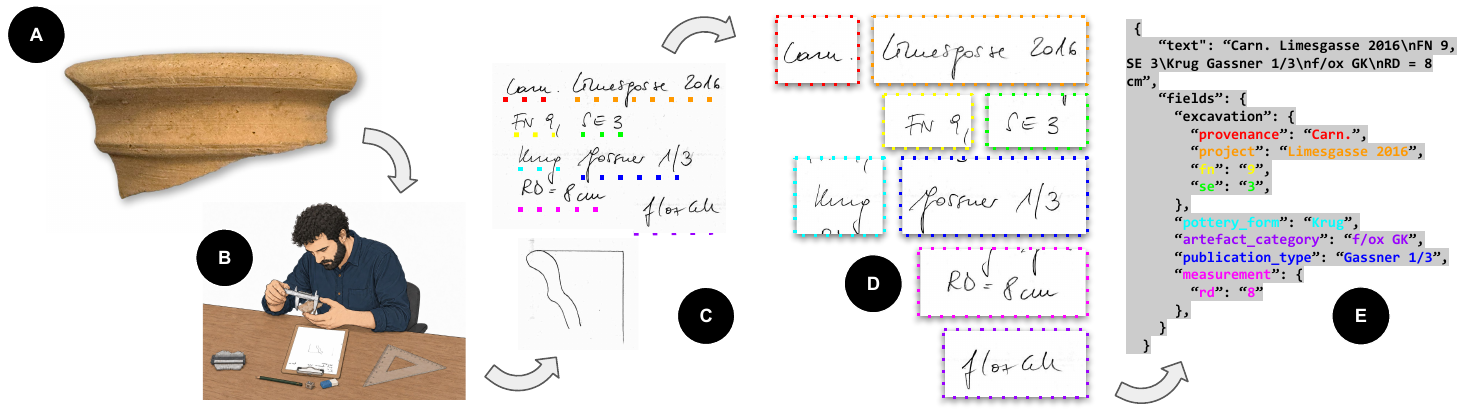} 
    \captionof{figure}{\textbf{From analogue archaeological documentation to structured metadata.} Ceramic sherds (A) are recorded by hand (B) and retro-digitised (C), yet handwritten annotations remain inaccessible to computational analysis. 
    We propose a pipeline for their automated conversion via HTR~(D) into machine-readable metadata~(E), and introduce the CENTURIA dataset to enable systematic benchmarking.}
    \label{fig:workflow}
\end{center}

\begin{abstract}    
    Pottery is a primary source for reconstructing the chronological and economic dimensions of past societies. Archaeologists often document ceramic finds through technical drawings and handwritten metadata. This metadata is critical for dating, provenance attribution, and cross-site comparison, but remains inaccessible to computational analysis, requiring manual transcription of every record. We investigate whether state-of-the-art document analysis models can address this task, and introduce \textbf{CENTURIA}, a dataset of 507 pottery records from the Roman site of Carnuntum, providing transcriptions, bounding boxes, and structured field-level labels across seven metadata categories. Benchmarking five OCR models reveals a substantial domain gap: zero-shot transcription error reaches 15--32\,\% SpACER-M, far exceeding rates on printed archival documents, with domain-specific fields recovered in fewer than 3\,\% of cases. LoRA fine-tuning on just 57 samples, reflecting a realistic archival annotation budget, closes this gap, reducing transcription error to below 1.5\,\%  and recovering overall field-level accuracy above 87\,\%. Our results show that a small expert-validated fine-tuning set suffices to convert handwritten pottery documentation into structured, searchable metadata ready for archaeological databases. 
\end{abstract}

\section{Introduction}

Archaeological archival collections comprise tens of thousands of technical 2D drawings of pottery sherds. Due to its durability and ubiquity, pottery is a primary source for reconstructing past societies~\cite{orton_pottery_2013}. Although digital recording methods are gaining importance, pottery documentation still largely follows traditional analogue procedures~\cite{zvietcovich2015assessment, pendic_use_2024,demjan_laser-aided_2023}: selected diagnostic fragments are drawn by hand as technical cross-sections, following comparatively standardised conventions that support comparison across sites and periods~\cite{buechner-matthews_drawing_2022}.

Even when these materials are retro-digitised, the annotations accompanying them remain inaccessible to computational analysis (\cref{fig:workflow} (A--C)): digitisation preserves only their visual appearance. These handwritten annotations are the primary carriers of excavation context, typological classification, and inventory numbers, indispensable for cross-site comparison, chronological reconstruction, and provenance attribution, yet their recovery requires intensive manual transcription, making archive-scale queries infeasible in practice~\cite{Hand2020}.

Handwritten Text Recognition (HTR) offers a practical path to making such archives machine-readable: approaches span from two-stage detection-recognition pipelines~\cite{li2021trocr} to full-page Vision-Language Models (VLMs)~\cite{olmocrbench} and lightweight specialised models~\cite{taghadouini2026lightonocr}, achieving character error rates below 5\,\% on standard handwritten benchmarks~\cite{garrido2025survey,crosilla2025benchmarking}.
However, it remains unclear how these advances transfer to archaeological material, as widely used benchmarks such as IAM~\cite{marti2002iam} and READ~\cite{sanchez2016read} cover only sequential text and do not reflect archival conditions, including spatially scattered annotations without a canonical reading order, faded ink, and dense domain-specific abbreviations.

In addition, raw transcriptions are insufficient for archive-scale analysis: retrieving vessels by form, filtering by ware type, or aggregating measurements requires structured metadata. As \cref{fig:workflow} (D--E) shows, Key Information Extraction (KIE)~\cite{huang2019sroie,rombach2025kie} bridges this gap by mapping transcribed text onto predefined semantic fields, where character-level errors compound into degraded extraction quality, making automated error detection crucial for practical deployment.

In this paper, we investigate whether state-of-the-art models can automate handwritten archaeological pottery documentation, with the practical aim of enabling heritage practitioners to make informed decisions about deploying these tools on their own archives. We show that zero-shot HTR models fall short, with transcription errors between 15 and 32\,\%, but LoRA fine-tuning on 57 samples, representative of a constrained annotation setting, reduces this to below 1.5\,\%, recovering overall field-level accuracy above 87\,\%. Our contributions are:

\begin{itemize}
    \item We introduce \textbf{CENTURIA}: the first benchmark for HTR and field extraction of handwritten archaeological records with scattered annotations and technical terminology, comprising 507 expert-validated samples.

    \item We present a systematic evaluation across five OCR models, three adaptation strategies, and two field extraction methods, assessing transcription quality, per-field performance, and error modes, showing that transcription quality is the primary determinant of field-level accuracy.

    \item We provide an end-to-end metadata extraction pipeline that closes the zero-shot domain gap with 57 fine-tuning samples, representative of realistic archival annotation budgets. By cross-combining models and extractors, we establish a heuristic agreement criterion that auto-accepts 56.4\,\% of scans at 95.5\,\% field precision, rising to 98.7\,\% on fields confirmed by all four pipelines, concentrating manual review on unresolved conflicts.
\end{itemize}
The complete CENTURIA dataset, fine-tuned checkpoints, and full code are released under CC-BY-4.0 at \url{https://github.com/gissuvalentina/CENTURIA}.
\section{Related Work}

\paragraph{Document Analysis in Archaeological Research.}
Archaeology increasingly draws on automated methods for visual recognition, classification, and 3D reconstruction of artefacts and sites~\cite{Bellat2025,DiAngelo2022,Gattiglia2025,Orengo2026}. For pottery specifically, prior work addresses the visual layer: ArchAIDE~\cite{anichini2020archaide} identifies ceramic types from photographs using shape- and appearance-based neural networks, while AutArch~\cite{klein2025autarch} detects objects and extracts geometric data from printed archaeological catalogues. For the archival text layer, OCR-based workflows produce structured metadata from printed heritage documents~\cite{bergamaschi2022digitalmaktaba}, and digital repositories such as tDAR~\cite{mcmanamon2017tdar} support access to and long-term preservation of digitised archaeological records. However, VLM-based generation of catalogue descriptions for archival photographic collections shows that hallucinations and domain-specific terminology make fully automated processing unreliable even at this level~\cite{abele2025archivegpt}. The challenge is more acute for handwritten records: recent attempts to digitise handwritten archaeological excavation notes confirm that training OCR on such material is time-intensive with low out-of-the-box accuracy~\cite{fletcher2023legacydata}. None of these approaches provides a solution for the handwritten annotation layer of archaeological records. The closest work to ours is the PyPottery toolkit~\cite{cardarelli2025pypotterylens,cardarelli2025pypotteryink,pypottery_toolkit_gh}, an open-source suite for the digitisation of archaeological pottery documentation. As part of its transcription component, PyPotteryScan~\cite{pypottery} applies OCR models to text annotations, but requires manual bounding box annotation as input and, to the best of our knowledge, reports no published evaluation of transcription quality or field extraction accuracy. We employ PyPotteryScan to generate preliminary transcriptions for the creation of ground truth; however, all documents require manual correction. Our work goes beyond PyPotteryScan by providing a fully automatic pipeline and the first systematic evaluation for handwritten pottery documentation.

\paragraph{Text Recognition Models and Field Extraction.}
HTR models evolved from line-level recognition to end-to-end full-page processing~\cite{garrido2025survey}: architectures range from specialised HTR systems~\cite{puigcerver2018pylaia,transkribus} and two-stage detection-recognition pipelines~\cite{li2021trocr} to full-page VLMs~\cite{olmocrbench,olmocr2}, generalist vision models~\cite{xiao2023florence2}, and compact reinforcement learning-optimised systems~\cite{taghadouini2026lightonocr}. Large pre-trained models can match specialised HTR engines on historical scripts~\cite{romein2025assessing,crosilla2025benchmarking}, and fine-tuning on small domain-specific samples substantially reduces error rates~\cite{kohut2023finetuning,loravsfullfinetuning}. Raw transcriptions alone, however, are insufficient for archive-scale analysis: KIE~\cite{huang2019sroie,rombach2025kie} addresses this by mapping text onto structured schemas using layout-aware~\cite{huang2022layoutlmv3} and OCR-free architectures~\cite{kim2022donut}, but existing datasets~\cite{simsa2023docile} assume printed business documents with standard layouts, neither of which holds for handwritten archaeological records. Transferability of these models and extraction methods to handwritten archaeological archives with scattered annotations and no canonical reading order remains unaddressed; we provide such an assessment to inform deployment decisions for heritage practitioners.

\paragraph{Benchmarks for HTR in Historical and Cultural Heritage Documents.} 
No existing benchmark covers archaeological documentation: systematic evaluation of HTR and KIE on historical documents centres on administrative and ecclesiastical records with well-defined structures. ESPOSALLES/IEHHR~\cite{romero2013esposalles,fornes2017icdar2017} provides semantic labels for handwritten marriage records, SIMARA~\cite{tarride2023simara} targets key-value extraction from archival index cards, and POPP~\cite{constum2022recognition} covers population census tables. Medieval corpora such as CATMuS Medieval~\cite{clerice2024catmus} address abbreviation-rich writing but focus on transcription rather than structured extraction. Record types and vocabulary across these resources are drawn from civil administration, genealogy, and ecclesiastical archives; archaeological field documentation, with its domain-specific abbreviations, heterogeneous record structures, and scattered annotations without a canonical reading order, is not represented. CENTURIA is the first benchmark assessing transcription and field-level KIE for this record type, covering model comparison, adaptation strategies, and extraction methods. 

\begin{figure}[tbp]
    \centering
    \includegraphics[width=\linewidth]{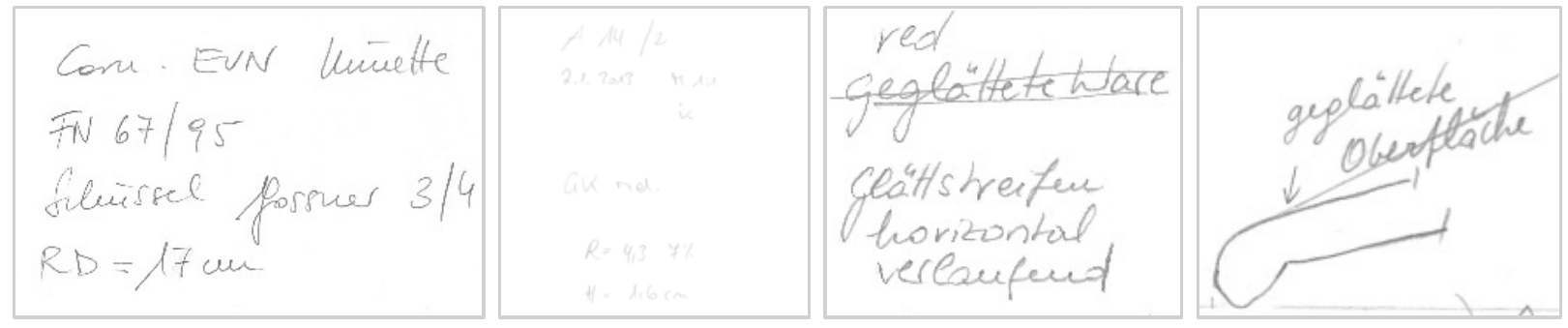}
\caption{\textbf{CENTURIA document crops.} From left to right: high-quality scan, low-contrast scan, crossed-out text, and handwriting overlapping with a technical drawing.}
\label{fig:sample_diversity}
\end{figure}

\setcounter{footnote}{0}

\section{CENTURIA Dataset}

CENTURIA comprises 507 annotated scans of analogue pottery records from the archaeological site of Carnuntum\footnote{\url{https://www.carnuntum.at/en}, last accessed 12 August 2026.}, the largest Roman settlement in present-day Austria and part of the UNESCO World Heritage Site ``Danube~Limes''~\cite{GuglEtAl2024}. Each scan is annotated with bounding boxes for text regions and technical drawings, manually transcribed text, and structured field-level values. A predefined train/test split (train: $n=57$, test: $n=450$), drawn by proportional stratified random sampling across six campaigns, supports reproducible evaluation and fine-tuning. 

\subsection{Data Collection}

CENTURIA is compiled from the Carnuntum documentation archive at the Austrian Archaeological Institute (OeAI)~\cite[for selected data]{GuglVadeanu2023}. The archive comprises approximately 70{,}000 analogue pottery records from which CENTURIA samples 507 documents across six campaigns from 1976--2017. Each record documents a common-ware pottery sherd predominantly in abbreviated German, and exhibits variation in handwriting style, scribal conventions, and document condition across campaigns and authors. Sampling is stratified to capture text diversity (handwriting style, character types, and annotation density across campaigns) and special cases (crossed-out text and overlapping annotations; illustrated in \cref{fig:sample_diversity}, with statistics in \Cref{tab:dataset_characteristics}). Appendix~\ref{app:dataset} lists the campaigns with their record counts and describes the sampling strategy in full.

\begin{table}[t]
\small 
\renewcommand{\arraystretch}{1}
\setlength{\tabcolsep}{4pt}
\centering
\caption{\textbf{CENTURIA dataset overview.} Dataset statistics, annotation effort, and stratification criteria; the latter reported as document counts and percentage share.} \label{tab:dataset_characteristics}
\begin{tabularx}{\textwidth}{@{} X r @{\hspace{2em}} X r r @{}}
\toprule
\multicolumn{2}{c}{\textbf{Dataset Statistics}} & \multicolumn{3}{c}{\textbf{Stratification Criteria}} \\
\cmidrule(lr){1-2} \cmidrule(l){3-5}
Total documents & 507 & \textbf{Annotation Density} & & \\
Total bounding boxes & 3345 & \quad Low (1--5 bbox) & 155 & 30.57\% \\
Total characters & 33767 & \quad Medium (6--8 bbox) & 291 & 57.40\% \\
Total words & 8261 & \quad High (9+ bbox) & 61 & 12.03\% \\
\textbf{Annotation Statistics} & & \textbf{Special Cases} & & \\
\quad Correction (manual) & 507 (100\%) & \quad Crossed-out text & 15 & 2.96\% \\
\quad Expert consultation & 48 (9.47\%) & \quad Overlapping annot. & 60 & 11.83\% \\
\bottomrule
\end{tabularx}
\end{table}

\begin{table}[t]
\centering
\scriptsize
\caption{\textbf{Metadata fields in CENTURIA with representative values.} The schema covers seven semantic categories; all fields are optional.} \label{tab:field_examples}
\renewcommand{\arraystretch}{1.15}
\begin{tabularx}{\textwidth}{@{}llX@{}}
\toprule
\textbf{Field} & \textbf{Subfield} & \textbf{Example Values} \\
\midrule
\multirow{7}{*}{\texttt{excavation}}
  & \texttt{provenance} & \textit{Carn.}, \textit{Car.}, \textit{CAR} \\
  & \texttt{project}    & \textit{GS 2012}, \textit{Survey GB 2017}, \textit{Limesgasse 2016} \\
  & \texttt{fn}         & \textit{206}, \textit{A\,7/1}, \textit{67/95} \\
  & \texttt{se}         & \textit{109}, \textit{142} \\
  & \texttt{fl}         & \textit{1}, \textit{2} \\
  & \texttt{quadrant}   & \textit{Q\,01--10}, \textit{Q\,44--02} \\
  & \texttt{kiste}      & \textit{Ki\,1/76}, \textit{Ki\,131/76} \\
\midrule
\texttt{pottery\_form}      & -- & \textit{Topf}, \textit{Schüssel}, \textit{Krug}, \textit{Deckel}, \textit{Amphore} \\
\texttt{artefact\_category} & -- & \textit{f/ox GK}, \textit{g/red GK}, \textit{f/ox PGW}, \textit{GK Mischbrand} \\
\texttt{publication\_type}  & -- & \textit{Gassner 3/4}, \textit{Petznek 17.6}, \textit{Grünewald 1979 Taf.\,54/14} \\
\texttt{surface\_treatment} & -- & \textit{Üz}, \textit{Üz\,a\,+\,i}, \textit{Glasur}, \textit{Grießbewurf} \\
\midrule
\multirow{3}{*}{\texttt{measurement}}
  & \texttt{rd}, \texttt{bd}                    & \textit{16\,cm}, \textit{5\,cm} \\
  & \texttt{h}                                  & \textit{2.1\,cm} \\
  & \texttt{r\_a}, \texttt{r\_i}, \texttt{r\_m} & \textit{8\,cm}, \textit{8\,cm\,(5\,\%)} \\
\midrule
\texttt{others} & -- & \textit{SPA}, \textit{M\,1:1}, \textit{14.12.2012}, \textit{Gez.\,MBC} \\
\bottomrule
\end{tabularx}
\end{table}

\subsection{Annotation Methodology}

\subsubsection{Transcription Annotation.} 
Bounding boxes for text regions and technical drawings are manually drawn following the annotation protocol detailed in Appendix~\ref{app:annotation}. To generate preliminary transcriptions, we use PyPotteryScan~\cite{pypottery}, an open-source tool built on olmOCR2~\cite{olmocr2} for semi-automatic transcription of archaeological pottery documentation. All 507 transcribed documents require manual correction, taking an average of 70 seconds per record. In 48 cases (9.47\,\% of documents), the correct reading cannot be determined from the scan alone and is resolved by consulting an expert in the archaeological domain. These cases cover illegible characters, ambiguous abbreviations, and site-specific terminology.

\subsubsection{Field Annotation.} 
\label{sec:field_annotation_dataset}

Each transcription is mapped onto a structured schema of seven semantic categories (illustrated in \Cref{tab:field_examples}), for example, \texttt{pottery\_form}: \textit{Krug} (jug), \texttt{artefact\_category}: \textit{f/ox GK} (oxidation-fired coarse ware). Full field descriptions are in Appendix~\ref{app:annotation}. To reduce manual annotation, a rule-based script applies regular expressions (REGEX) and controlled vocabularies to the corrected transcriptions, matching and assigning text spans to their corresponding fields. Unrecognised text is assigned to an \textit{others} category. The resulting field annotations are verified by a Carnuntum site specialist, with corrections applied in 3.75\,\% of the documents, all involving edge cases such as crossed-out entries, field boundary errors, and ambiguous plate references (\textit{Taf.}~X).

\section{Method}

\begin{figure}[t]
    \centering
    \includegraphics[width=\linewidth]{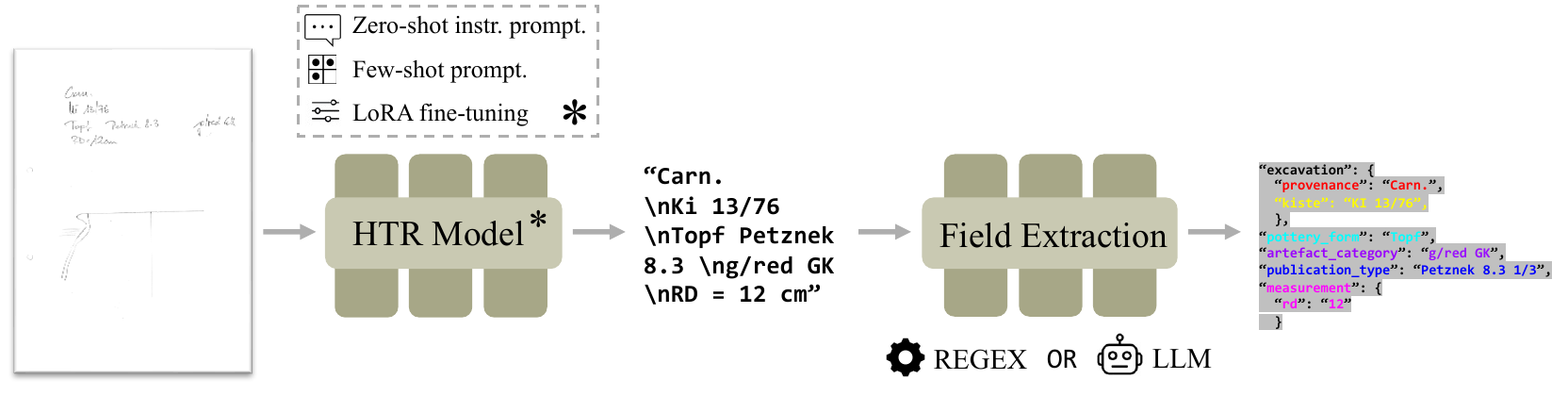}
    \caption{\textbf{Two-stage pipeline: HTR followed by structured field extraction.} A scanned pottery record is transcribed by an HTR model (optionally adapted via instruction prompting, few-shot prompting, or LoRA fine-tuning), then parsed into structured metadata fields via REGEX or LLM-based (Qwen3-8B) extraction.}
    \label{fig:method_pipeline}
\end{figure}

Given a scanned pottery documentation record, the method produces structured metadata fields through two stages: text recognition and field extraction (\cref{fig:method_pipeline}).

\subsection{Handwritten Text Recognition}
Each OCR model $f_\theta$ takes a single-page scan $x$ as input and produces a character sequence $\hat{y} = f_\theta(x)$ as a transcription of the handwritten content. Raw model output undergoes post-processing $\tilde{y} = g(\hat{y})$ to normalise formatting artefacts before field extraction (post-processing steps are described in Appendix~\ref{app:implementation}).

\subsubsection{Baseline Models.}
The models span the following paradigms: a traditional two-stage detection-recognition pipeline (\textit{TrOCR}), first- and second-generation full-page VLMs (\textit{olmOCR}, \textit{olmOCR2}), a generalist multi-task model (\textit{Florence2}), and a lightweight reinforcement-learning-optimised model (\textit{LightOnOCR}). 
All models are open-source, prioritising reproducible and cost-free deployment for heritage practitioners; checkpoints are detailed in Appendix~\ref{app:checkpoints}.

\paragraph{Two-stage detection-recognition.} 
\textit{TrOCR}~\cite{li2021trocr} pairs a Vision Transformer (ViT) encoder with a language model decoder pre-trained on large-scale text data. The encoder splits the detected text into image patch sequences and the decoder generates the transcriptions at wordpiece level conditioned on these embeddings. As TrOCR requires pre-segmented text regions, we use CRAFT~\cite{baek2019characterregionawarenesstext} via the EasyOCR library~\cite{easyocr} for line segmentation. Unlike full-page models, each region is processed independently without access to the document context, and transcription quality therefore depends directly on detection. The output consists of recognised text sequences with corresponding bounding box coordinates. 

\paragraph{Full-page vision-language models.} 

Unlike \textit{TrOCR}, these models take the entire document image as input, bypassing explicit text detection. \textit{olmOCR}~\cite{olmocrbench} represents the first generation of full-page VLMs for OCR, built on Qwen2-VL-7B~\cite{Qwen2VL} and trained for structured document understanding across complex layouts.
\textit{olmOCR2}~\cite{olmocr2} builds on Qwen2.5-VL-7B-Instruct~\cite{qwen2.5-VL}, the successor of Qwen2-VL-7B, which extends its predecessor with dynamic resolution processing and finer-grained document parsing. It further specialises for OCR via reinforcement learning with verifiable rewards (RLVR) on synthetic unit tests, improving over \textit{olmOCR} on structured document layouts. \textit{Florence2}~\cite{xiao2023florence2} adopts a unified sequence-to-sequence architecture across vision tasks and is included to examine whether broad multi-task pre-training can compete with specialised approaches on handwritten domain-specific documents. 

\paragraph{Lightweight specialised model.} 
\textit{LightOnOCR}~\cite{taghadouini2026lightonocr, lightonocr2_hf} is a 1B-parameter end-to-end VLM that takes a full image as input and outputs a text transcription. It consists of a native resolution ViT encoder initialised from Mistral-Small-3.1~\cite{mistral3}, a two-layer MLP projector that reduces visual token count via spatial patch merging, and a Qwen3~\cite{qwen3technicalreport} decoder that generates the transcription conditioned on the projected visual tokens without task prompts at inference. The model is pretrained end-to-end by distillation from a larger VLM teacher and post-trained with RLVR. At 1B parameters, LightOnOCR is roughly an order of magnitude smaller than the strongest full-page VLMs on olmOCR-Bench~\cite{taghadouini2026lightonocr} and represents the lightweight end of the baseline model spectrum.

\subsubsection{Domain Adaptation Strategies.}
\label{sec:adaptation}

Pre-training covers general document OCR, with the exception of TrOCR, which uses a checkpoint specifically fine-tuned on the IAM handwriting dataset~\cite{marti2002iam,trocr_large_handwritten_hf}. Nevertheless, a domain gap in archaeological vocabulary and annotation conventions limits zero-shot transcription quality across all models. We therefore explore three adaptation strategies: zero-shot instruction prompting~\cite{crosilla2025benchmarking}, few-shot prompting~\cite{wolf2025cm1}, and LoRA fine-tuning~\cite{hu2021lora}.

\paragraph{Zero-shot instruction prompting} augments the model input with a domain-specific instruction prompt $c$, conditioning the model as $\hat{y} = f_\theta(x, c)$. Since the models are not exposed to archaeological terminology during training, explicit instructions aim to improve recognition of domain-specific abbreviations and notation without requiring labelled data. The full prompt is used for all models; design details are in Appendix~\ref{app:prompt:zeroshot} and the complete prompt is in Appendix~\ref{app:prompts}.

\paragraph{Few-shot prompting} extends this by providing $k$ annotated examples $\{(x_i, y_i)\}_{i=1}^{k}$ alongside the input, conditioning the model as $\hat{y} = f_\theta(x \mid (x_1, y_1), \ldots, (x_k, y_k))$. Demonstrations of the expected input-output format allow the model to infer transcription conventions and domain vocabulary without weight updates. We use $k \in \{1, 5, 8\}$ examples. For reproducibility, we provide full prompt templates and example selection details in Appendices~\ref{app:prompt:fewshot} and~\ref{app:prompts}.

\paragraph{LoRA fine-tuning~\cite{hu2021lora}} adapts model weights on the CENTURIA training set ($n=57$) using parameter-efficient fine-tuning via low-rank weight decomposition $\Delta W = BA$, where $B \in \mathbb{R}^{d_{\mathrm{out}} \times r}$ and $A \in \mathbb{R}^{r \times d_{\mathrm{in}}}$ with rank $r \ll \min(d_{\mathrm{in}}, d_{\mathrm{out}})$, applied to all linear projection layers. Hyperparameters are given in Appendix~\ref{app:lora}.

\subsection{Field Extraction} 

Field extraction (KIE~\cite{huang2019sroie,rombach2025kie}) maps a transcription $\tilde{y}$ onto a predefined schema of $N=7$ semantic categories $\mathcal{K} = \{k_1, \ldots, k_N\}$, producing field-value pairs $\mathcal{F} = \{(k_i, v_i)\}_{i=1}^{N}$, where $v_i \in \mathcal{V}_{k_i} \cup \{\texttt{null}\}$ and $\mathcal{V}_{k_i}$ is the value domain for field $k_i$. Since field boundaries are not marked in the original records, the mapping $\tilde{y} \rightarrow \mathcal{F}$ relies entirely on textual patterns and domain vocabulary. We compare two extraction methods that answer different questions: a rule-based approach relying on exact pattern matching, which establishes the accuracy attainable under transcription noise, and an LLM-based approach operating on the full transcription sequence, which tests whether extraction transfers without archive-specific engineering.

\paragraph{REGEX-based extraction} applies the same rule-based script used to produce ground truth metadata field annotations (\cref{sec:field_annotation_dataset}), using regular expressions and controlled vocabularies $\mathcal{V} = \{V_k\}_{k=1}^{N}$ derived from the field schema, such that $v_i = \text{match}(\tilde{y}, V_{k_i})$ for categorical fields and $v_i = \text{regex}(\tilde{y}, p_{k_i})$ for open-domain fields such as measurements, where $p_{k_i}$ is the pattern for field $k_i$. To prevent duplicate assignments, matched substrings are immediately removed from the prediction; remaining unmatched text is aggregated into \textit{others}. Sharing the ground truth rule set, REGEX is correct by construction on verified transcriptions and fails only on OCR noise. It therefore isolates noise propagation and shows what deterministic rules achieve once transcription is accurate, without inference hardware but with archive-specific patterns. 

\paragraph{LLM-based extraction} uses an unmodified Qwen3-8B \cite{qwen3technicalreport} to approximate $\tilde{y} \rightarrow \mathcal{F}$ by prompting the model with cleaned transcription $\tilde{y}$ and the field schema $\mathcal{K}$, returning a structured JSON object with field-value pairs. The model is prompted in an 8-shot setting, with examples pairing corrected transcriptions with their expected field annotations; corrected examples perform slightly better than corrupted ones, particularly on publication type (Appendix~\ref{app:effect_fewshot_quality}). Fields absent or unrecognisable in the transcription are assigned \texttt{null}; tokens that do not match any schema field are collected in the \texttt{others} list. Unlike REGEX, the LLM uses a natural language prompt rather than authored patterns, which simplifies adaptation and lets it infer fields from corrupted tokens. For reproducibility, further details and the complete prompt are provided in Appendices~\ref{app:prompt:llm} and~\ref{app:prompts}.

\section{Evaluation}

We evaluate all five models on transcription quality, and the two strongest baselines (olmOCR2 and LightOnOCR) on domain adaptation, field extraction, and cross-pipeline confidence estimation. Processing time is recorded to assess scalability. Full results, qualitative examples, details of error analysis, and confidence estimation are provided in  Appendices~\ref{app:field_results} to \ref{app:inference_speed} and \ref{app:smart_merge}.

\subsection{Evaluation Metrics}
\subsubsection{Transcription.}
Following~\cite{bourne2026charactererrorvector}, we evaluate transcription quality using SpACER, as CER assumes a fixed linear reading order that does not hold for scattered annotations in archaeological ceramic records. For models providing bounding boxes (TrOCR, Florence2), Micro SpACER (SpACER-m) anchors predictions to ground truth via IoU matching; for VLMs without spatial outputs, Macro SpACER (SpACER-M) pools characters page-wide:
$$\text{SpACER-m} = \frac{\sum^n_j D_j + \hat{E}}{2C}, \qquad \text{SpACER-M} = \frac{D + \hat{E}}{2C}$$

where $D_j$ are deletions in matched region $j$, $D = \max(0, |g| - |p|)$ is the page-level deletion count, $g$ and $p$ are the ground truth and predicted character count vectors, $\hat{E} = \|g - p\|_1$ is the global character count difference, and $C = |g|$ is the total ground-truth character count.
Following~\cite{strobel2020much}, we additionally report BoW-F1, computed as $F1 = 2 \cdot \frac{|P \cap G|}{|P| + |G|}$ over the multi-set intersection of predicted and ground truth word sets, evaluating keyword presence regardless of position.

\subsubsection{Field-Level Transcription.}

To isolate transcription errors from extraction errors and identify which field types are most affected by OCR noise, we use Exact Match Rate (EMR) and Average Normalised Levenshtein Similarity (ANLS), following~\cite{9011031}. EMR checks whether $v_i^*$ appears exactly in $\tilde{y}$, computed per-document and per-field type. Since a single misread character yields EMR~$=0$, ANLS complements it with a normalised edit-distance similarity between $\hat{v}_i$ and $v_i^*$ ($\tau = 0.5$), tolerating minor OCR errors while assigning zero similarity to predictions that differ by more than half their characters.

\subsubsection{Field Extraction.}

Predicted field values $\hat{\mathcal{F}}$ are matched against ground truth annotations $\mathcal{F}^*$ after string normalisation, including lowercasing, whitespace collapsing, and delimiter standardisation. Following~\cite{varex2026}, we report Per-Field Accuracy and macro-averaged Overall Accuracy, ignoring unannotated fields.

\subsection{Transcription Results}

\subsubsection{Baseline Performance.} 
\begin{figure}[tb]
    \centering
    \begin{minipage}[b]{0.48\linewidth}
        \centering
        \includegraphics[width=\linewidth]{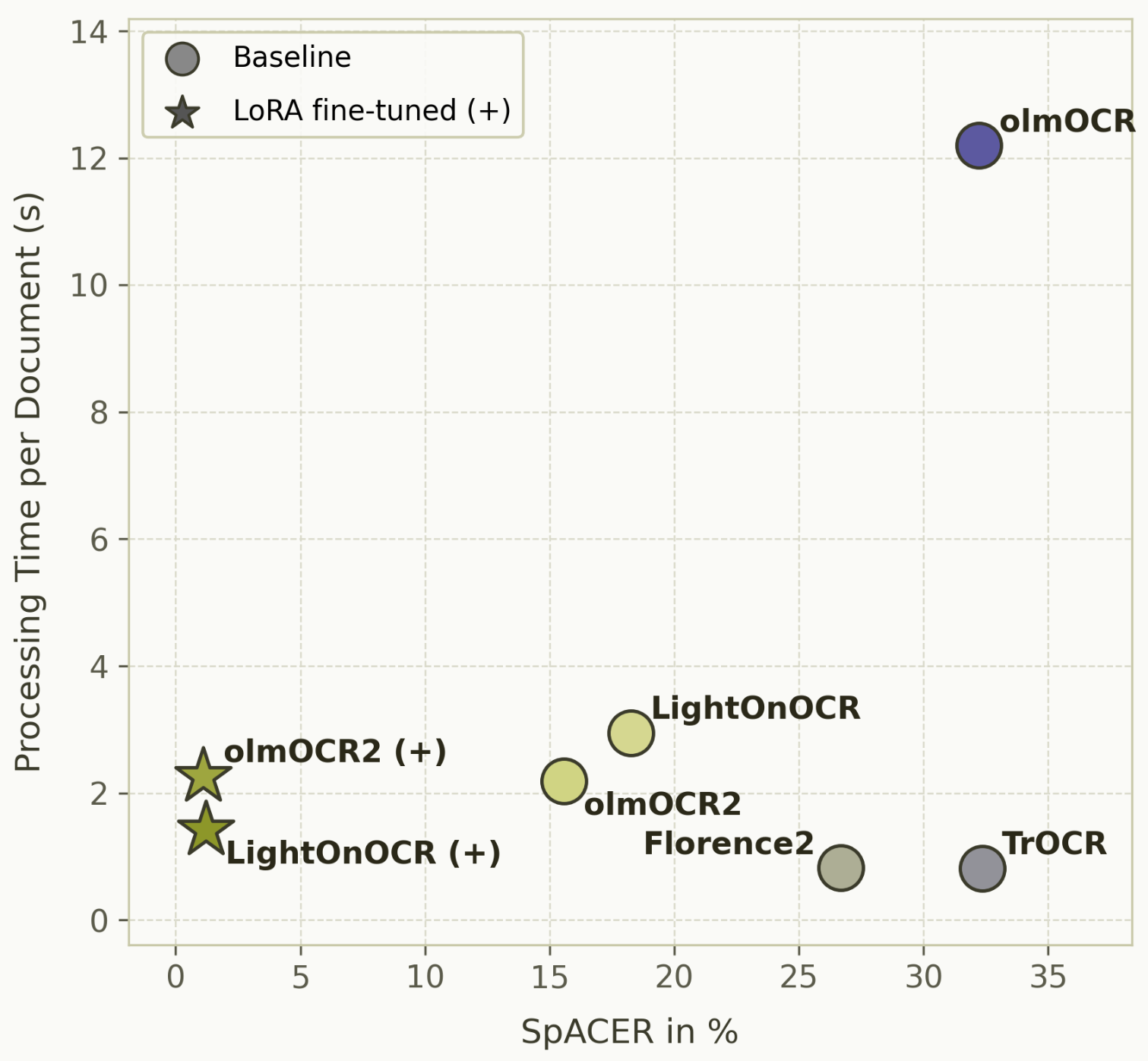}
        \caption{\textbf{Transcription error vs.\ processing time.} LoRA fine-tuned models ($\bigstar$) achieve the best accuracy-efficiency trade-off, with SpACER-M around 1\,\%.}
        \label{fig:ocr_efficiency}
    \end{minipage}
    \hfill
    \begin{minipage}[b]{0.48\linewidth}
        \centering
        \includegraphics[width=\linewidth]{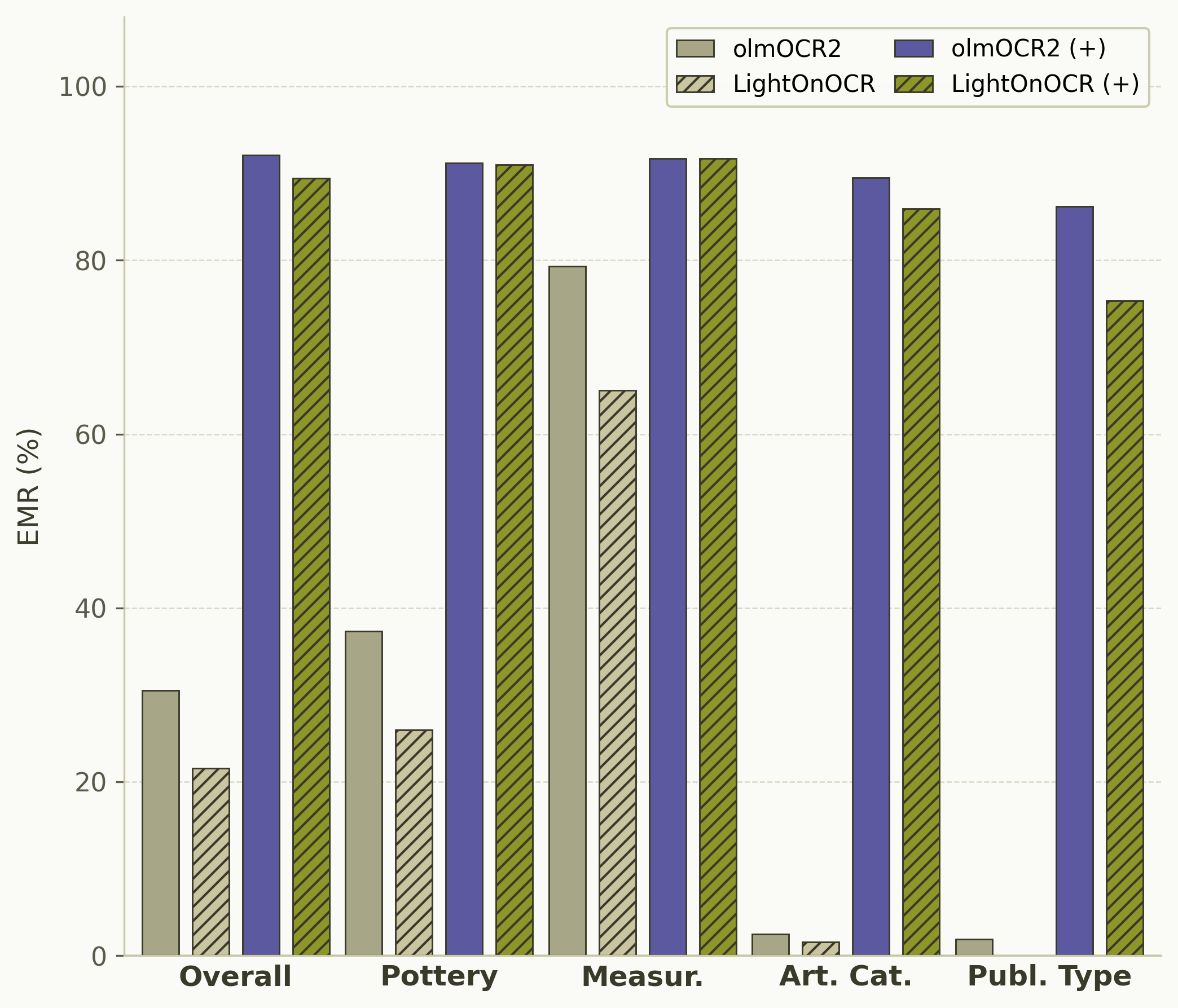}
    \caption{\textbf{Field-level EMR by category.} Baseline models fail on domain-specific fields; LoRA fine-tuning achieves $\sim90\,\%$ overall EMR.}
    \label{fig:field_emr}
    \end{minipage}
\end{figure}

\Cref{tab:results} shows that zero-shot transcription is challenging across all models (SpACER-M 15.58--32.35\,\%), exceeding the 10.6\,\% reported for olmOCR on printed archival documents~\cite{bourne2026charactererrorvector} and confirming the difficulty of handwritten pottery documentation. Field-level accuracy is similarly low: even the strongest baseline, olmOCR2, achieves only 30.55\,\% EMR and 57.43\,\% ANLS, indicating that field values are rarely recovered verbatim from zero-shot output. \Cref{fig:ocr_efficiency} illustrates the accuracy-efficiency trade-off: TrOCR and Florence2 are the fastest models (under 1\,s/doc) but among the least accurate, while olmOCR is the least efficient (12.2\,s/doc) without a corresponding accuracy advantage. We therefore restrict subsequent domain adaptation and field extraction to the two most accurate baselines, olmOCR2 and LightOnOCR. 

\begin{table}[t]
\centering
\scriptsize
\setlength{\tabcolsep}{8pt}
\caption{\textbf{Character and word-level OCR performance with field-level accuracy and computational efficiency.} SpACER-m only reported for TrOCR and Florence2. EMR and ANLS are per-document averages across all GT fields (test set, $n=450$). Timing measured on NVIDIA GeForce RTX 3090 ($2 \times 24$\,GB VRAM). $(+)$ denotes models fine-tuned with LoRA ($n=57$). Best in \textbf{bold}, second best \underline{underlined}.}
\label{tab:results}
\begin{tabularx}{\textwidth}{@{}lZ{1.3}Z{1.3}Z{0.85}Z{0.85}Z{0.85}Z{0.85}@{}}
\toprule
\textbf{Model} & \textbf{\shortstack{SpACER-m\\ $\downarrow$ (\%)}} & \textbf{\shortstack{SpACER-M\\ $\downarrow$ (\%)}} & \textbf{\shortstack{BoW-F1\\ $\uparrow$ (\%)}} & \textbf{\shortstack{EMR\\ $\uparrow$ (\%)}} & \textbf{\shortstack{ANLS\\ $\uparrow$ (\%)}} & \textbf{\shortstack{t/Doc\\ $\downarrow$ (s)}} \\
\midrule
TrOCR        & 37.50 & 32.35 & 41.81 & 7.13 & 32.81 & 0.81 \\
olmOCR       & —     & 32.22 & 43.47 & 13.71 & 30.07 & 12.20 \\
olmOCR2      & —     & 15.58 & 56.74 & 30.55 & 57.43 & 2.19 \\
Florence2    & 31.82 & 26.67 & 37.69 & 17.39 & 51.13 & 0.83 \\
LightOnOCR   & —     & 18.25 & 50.98 & 21.59 & 47.62 & 2.95 \\
\midrule
olmOCR2~(+)     & — & \textbf{1.11}    & \textbf{96.24}  & \textbf{92.08}  & \textbf{95.29}  & 2.26 \\
LightOnOCR~(+)  & — & \underline{1.21} & \underline{94.80} & \underline{89.40} & \underline{94.75} & 1.42 \\
\bottomrule
\end{tabularx}
\end{table}

\subsubsection{Domain Adaptation.} 
\Cref{tab:improvement_strategies} evaluates prompt-based adaptation strategies on olmOCR2 and LightOnOCR; full results across all strategies and shot counts are in Appendix~\ref{app:improvement_strategies_full}. Zero-shot instruction prompting substantially improves olmOCR2 (SpACER-M: 15.58\,\%~$\to$~6.77\,\%, BoW-F1: 56.74\,\%~$\to$~83.08\,\%), while its effect on LightOnOCR is smaller. Few-shot prompting reduces olmOCR2 error (SpACER-M 5.38\,\% at 5 shots), but collapses LightOnOCR into repetitive output loops (SpACER-M 48.71\,\% at 1 shot, 100\,\% at 5 shots), consistent with known repetition failure modes~\cite{holtzman2019curious} and likely reflecting its RLVR training towards short, prompt-free outputs (Appendix~\ref{app:improvement_strategies_full}). LoRA fine-tuning nonetheless dominates all prompting strategies. With just 57 samples, SpACER-M drops from over 15\,\% to below 2\,\% for both models, showing that even minimal fine-tuning data outperforms prompt engineering alone. LightOnOCR~(+) additionally reduces inference time through shorter outputs, while olmOCR2~(+) shows no such gain (details in Appendix~\ref{app:inference_speed}). This confirms that effective domain adaptation requires only a small annotation effort, making the approach viable for archives without large labelled collections.

\begin{table}[t]
\centering
\scriptsize
\setlength{\tabcolsep}{6pt}
\caption{\textbf{Domain adaptation strategies.} Applied to olmOCR2 and LightOnOCR. Best in \textbf{bold}, second best \underline{underlined}.}
\label{tab:improvement_strategies}
\begin{tabularx}{\textwidth}{@{}llcccc@{}}
\toprule
\textbf{Model} & \textbf{Strategy} & \textbf{\shortstack{SpACER-M\\ $\downarrow$ (\%)}} & \textbf{\shortstack{BoW-F1\\ $\uparrow$ (\%)}} & \textbf{\shortstack{EMR\\ $\uparrow$ (\%)}} & \textbf{\shortstack{ANLS\\ $\uparrow$ (\%)}} \\
\midrule
\multirow{4}{*}{olmOCR2}
 & Baseline (zero-shot)              & 15.58 & 56.74 & 30.55 & 57.43 \\
 & Zero-shot instr. prompting        & 6.77 & 83.08 & 74.64 & 84.99 \\
 & Few-shot (5 shots)                & 5.38 & 84.44 & 68.97 & 87.51 \\
 & LoRA fine-tuning (2 epochs)       & \textbf{1.11} & \textbf{96.24} & \textbf{92.08} & \textbf{95.29} \\
\midrule
\multirow{4}{*}{LightOnOCR}
 & Baseline (zero-shot)              & 18.25 & 50.98 & 21.59 & 47.62 \\
 & Zero-shot instr. prompting        & 14.13 & 53.71 & 35.27 & 65.14 \\
 & Few-shot (1 shot)                 & 48.71 & 43.57 & 28.02 & 67.24 \\
 & LoRA fine-tuning (2 epochs)       & \underline{1.21} & \underline{94.80} & \underline{89.40} & \underline{94.75} \\
\bottomrule
\end{tabularx}
\end{table}

\subsubsection{Field-level Performance.}

\cref{fig:field_emr} shows field-level EMR for olmOCR2 and LightOnOCR; full results are in \Cref{tab:field_results} (Appendix~\ref{app:field_results}). Baseline models handle measurements well (EMR 65--80\,\%) as numeric values are visually robust. Pottery form shows intermediate performance, with baseline models recovering it in 26--37\,\% of cases. 
Domain-specific fields fail entirely: Artefact Category and Publication Type reach near-zero EMR as OCR noise corrupts abbreviations and domain tokens beyond recognition (e.g.\ \textit{f/ox\,GK}~$\to$~\textit{florak}). LoRA fine-tuning resolves this, with overall EMR exceeding 89\,\% for both models.

\subsection{Field-extraction Results}

\subsubsection{Extraction Accuracy.} 
\Cref{tab:field_extraction} reports field extraction accuracy under REGEX and LLM (Qwen3-8B). At baseline, performance is low and nearly identical regardless of extraction method (29--30\,\% for olmOCR2, 26--27\,\% for LightOnOCR), pointing to the shared OCR input as the limiting factor rather than the extraction method itself. LoRA fine-tuning confirms this: once transcription quality improves, extraction accuracy follows, with olmOCR2~(+) outperforming LightOnOCR~(+) (91.73\,\% vs.\ 88.30\,\% under REGEX) notably on Publication Type and Surface Treatment, with gaps of 14.91 and 20.37 percentage points.

\begin{table}[t]
\centering
\footnotesize
\caption{\textbf{Field extraction accuracy (\%) under REGEX and LLM extraction~(Qwen3-8B).} Overall is the per-document average across all fields. Excavation aggregates provenance, crate, project, FN, SE, FL, and quadrant (weighted by $n$). $(+)$ denotes LoRA fine-tuned models.
\textbf{Bold}: best per column.}
\label{tab:field_extraction}
\resizebox{\linewidth}{!}{%
\begin{tabular}{@{}lcccccc|cccccc@{}}
\toprule
& \multicolumn{6}{c}{\textbf{REGEX Extraction}}
& \multicolumn{6}{c}{\textbf{LLM Extraction (Qwen3-8B)}} \\
\cmidrule(lr){2-7} \cmidrule(lr){8-13}
\textbf{Model}
  & \textbf{Ovr.} & \textbf{Excav.} & \textbf{Form}
  & \textbf{Ware} & \textbf{Pub.} & \textbf{Surf.}
  & \textbf{Ovr.} & \textbf{Excav.} & \textbf{Form}
  & \textbf{Ware} & \textbf{Pub.} & \textbf{Surf.} \\
\midrule
olmOCR2
  & 29.96 & 31.33 & 35.50 &  3.79 & 0.54 & 0.00
  & 29.12 & 28.93 & 34.34 &  2.23 & 0.27 & 0.00 \\
LightOnOCR
  & 26.75 & 27.55 & 30.86 &  3.57 & 0.27 & 0.00
  & 26.61 & 25.92 & 28.31 &  1.79 & 0.00 & 0.00 \\
\midrule
olmOCR2~(+)
  & \textbf{91.73} & \textbf{96.05} & \textbf{90.02}
  & \textbf{90.18} & \textbf{85.91} & \textbf{61.11}
  & \textbf{90.76} & \textbf{95.96} & \textbf{88.86}
  & \textbf{88.84} & \textbf{82.11} & \textbf{57.41} \\
LightOnOCR~(+)
  & 88.30 & 94.08 & 89.56 & 87.28 & 71.00 & 40.74
  & 87.47 & 94.16 & \textbf{88.86} & 85.04 & 68.83 & 35.19 \\
\bottomrule
\end{tabular}}
\end{table}


Since REGEX strictly enforces the annotation convention, its primary failure mode is a character-level OCR error that breaks a match, whereas the LLM receives the same field convention as a prompt and can additionally deviate from it. Evaluating both extractors on ground truth transcriptions instead of OCR output separates the two effects: REGEX reaches 99.53\,\% overall accuracy, failing on the 16 expert-corrected records outside its rule coverage (\cref{sec:field_annotation_dataset}), while the LLM achieves 96.76\,\%. This demonstrates that the baseline gap of 2.77 percentage points arises independently of OCR noise and reflects the LLM's divergence from the annotation convention. On OCR output the two extractors differ by less than one point overall for both baseline and fine-tuned models, as OCR noise costs REGEX more than the LLM and partially offsets this deviation. The REGEX--LLM gap is most pronounced on Publication Type for olmOCR2~(+) (85.91\,\% vs.\ 82.11\,\%), where the LLM frequently extracts only one reference from multi-reference entries, even though the prompt explicitly allows list output. The corresponding error breakdown and model overlap analysis are provided in Appendix~\ref{app:field_extraction}.

\subsubsection{Error Modes.}
\begin{figure}[t]
  \centering
  \includegraphics[width=\linewidth]{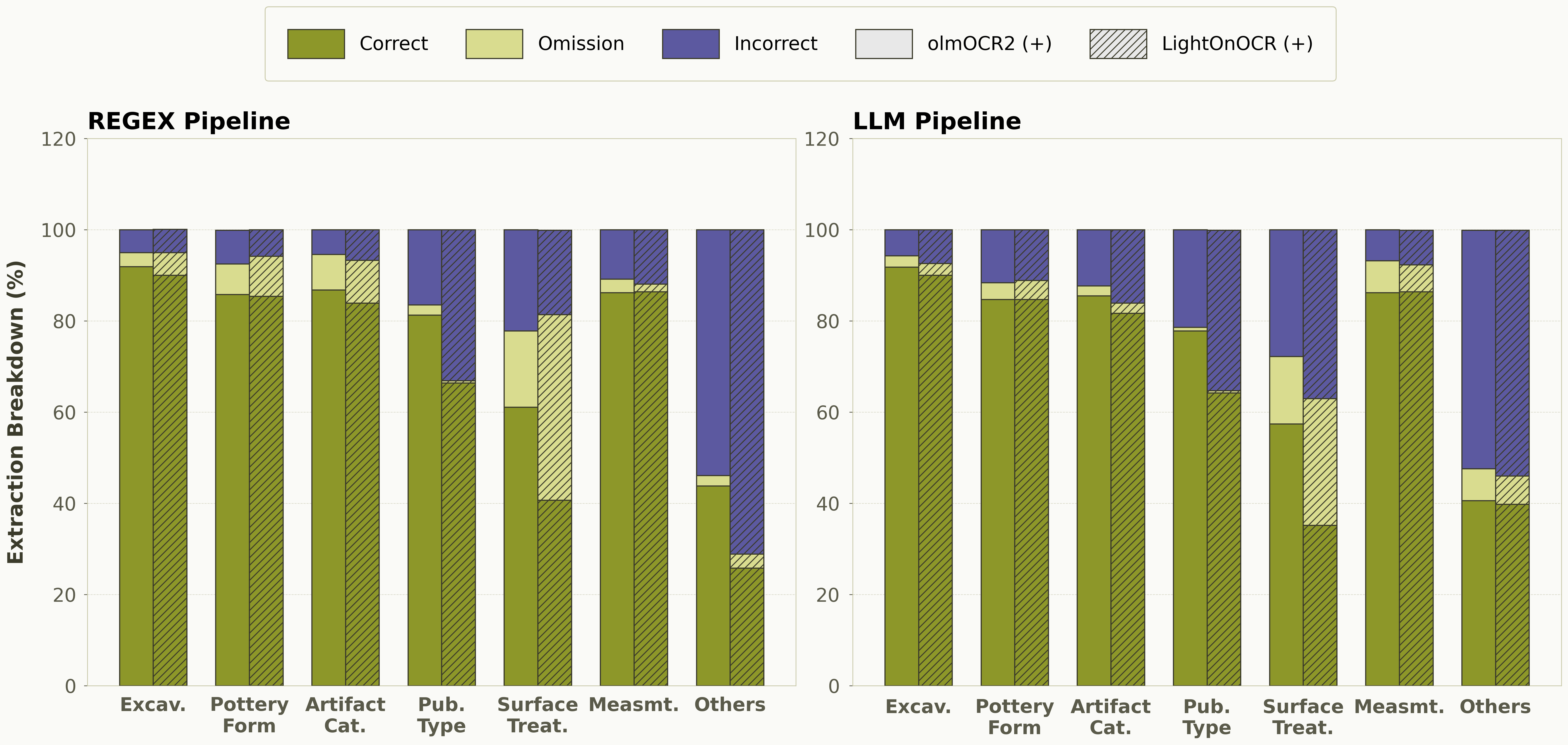}
  \caption{\textbf{Extraction performance across all fields.} Error breakdown (\%) for olmOCR2~(+) and LightOnOCR~(+) across the seven metadata categories, showing correct extractions, omissions, and incorrect predictions, under REGEX and LLM extraction.}
  \label{fig:extraction_both}
\end{figure}

\cref{fig:extraction_both} breaks down extraction outcomes per field into three categories: correct, omission (field absent in the output), and incorrect (field present but wrong value). Publication Type shows the highest incorrect rate under both pipelines, driven by digit-level OCR errors producing plausible but wrong values. Fields with short or domain-specific tokens (Pottery Form, Artefact Category, Surface Treatment) show higher omission rates under REGEX, as corrupted tokens fall outside pattern coverage; under LLM extraction, these cases shift towards incorrect predictions, as the LLM selects the closest vocabulary match rather than returning null. This shift is most pronounced for Surface Treatment, where LightOnOCR~(+) produces ${\sim}$41\,\% omissions under REGEX but ${\sim}$37\,\% incorrect predictions under LLM. Measurement shows notably higher omission under LLM than REGEX, suggesting the LLM more frequently fails to extract numeric values entirely. That the failures originate in OCR noise rather than extraction method choice is confirmed by model overlap (\Cref{tab:extraction_errors}, Appendix~\ref{app:field_extraction}): 60\,\% of REGEX errors and 71\,\% of LLM errors are shared by both models.

\subsubsection{Confidence Estimation.}
Cross-checking field predictions across two models and two extraction methods serves as a practical heuristic for archive-scale deployment. While overall field extraction accuracy already exceeds 87\,\%, fields on which all four pipeline combinations (olmOCR2~(+) and LightOnOCR~(+), each paired with REGEX and LLM extraction) fully agree are correct in 98.7\,\% of cases. Extending acceptance to partial agreement, resolved by majority vote and field-specific rules, balances conditional precision against broader coverage by auto-accepting 56.4\,\% of documents at an overall precision of 95.5\,\%. Therefore, manual review concentrates on the remaining scans, where disagreement persists. A preliminary confidence scheme for partial disagreements is detailed in Appendix~\ref{app:smart_merge}.

\section{Limitations and Future Work}
\label{app:limitations}

While CENTURIA and the proposed pipeline demonstrate promising results for automated processing of handwritten pottery records, several limitations remain.

\paragraph{Remaining failure modes.} Despite strong overall performance, systematic failure modes persist after fine-tuning. Short domain-specific tokens (e.g.\ surface treatment abbreviations) are prone to omissions under REGEX extraction, while digit-level errors in structured fields (excavation numbers, publication type) produce plausible but incorrect predictions across both pipelines. Percentage radius measurements (\texttt{r\_a}, \texttt{r\_i}) remain unresolved, as no pipeline reliably transcribes the \% sign. Additionally, struck-through text in the original records is not filtered during transcription. OCR models read both the correction and the crossed-out text, which can introduce noise into downstream field extraction.

\paragraph{Quality validation and human evaluation.} 
Cross-pipeline confidence estimation offers practical triage: fields confirmed by all four pipelines are correct in 98.7\,\% of cases across the test set, and the 56.4\,\% of documents that are auto-accepted reach 95.5\,\% precision. Both figures are conditional on agreement-selected subsets rather than the full test set, and the pipelines are not fully independent (shared LoRA training data; REGEX shares the ground truth rule set), so agreement may overestimate reliability. While we explore a preliminary active learning scheme (Appendix~\ref{app:smart_merge}), we plan to formalise this into a full human-in-the-loop workflow. Future work will additionally evaluate time savings and error correction effort with domain experts in real archival settings.

\paragraph{Transferability and scope.} As the first benchmark for HTR and field extraction on handwritten archaeological pottery documentation, CENTURIA opens the question of transferability across sites and documentation practices. The core challenges CENTURIA captures, namely handwritten abbreviations, spatially scattered annotations, and domain-specific vocabulary, are common to archaeological pottery documentation worldwide~\cite{wiegmann_zeichenrichtlinien_2018,collett_introduction_2017,macginnis_ceramic_2022,kosak_instruction_2023,okayama_board_education_2018,moreno_quixal_2013}. CENTURIA itself spans six campaigns with variation in handwriting style, scribal conventions, and document condition, providing evidence that the approach remains effective under within-archive diversity, though cross-archive generalisation additionally depends on the target archive's recording conventions, script, and language. The training split size of 57 is designed to reflect a realistic annotation budget in archival settings; how performance scales with annotation effort, and what minimum suffices for archives with different handwriting variation or recording conventions, remains an open question for future work. Cross-archive evaluation is non-trivial given restricted access to archaeological collections; we aim to extend CENTURIA to other sites where access permits. In the longer term, aggregating annotated records across multiple archives could support the development of a foundational OCR model for archaeological documentation, reducing the annotation burden for individual archives through shared pre-training.

\section{Conclusion}







We introduced CENTURIA, 507 annotated handwritten pottery records from the Carnuntum excavations, and used it to benchmark five OCR models on archaeological HTR and structured field extraction. On CENTURIA, transcription quality rather than extraction method governs end-to-end performance: zero-shot models fail (15--32\,\% SpACER-M), but LoRA fine-tuning on only 57 samples reduces transcription error below 1.5\,\% and lifts overall field accuracy above 87\,\%, after which structured metadata recovery follows from either extraction method (REGEX or LLM-based). Two further results make the approach practical for archival deployment. LightOnOCR~(+) achieves comparable overall accuracy to olmOCR2~(+) at substantially lower computational cost, and cross-pipeline agreement auto-accepts 56.4\,\% of documents at 95.5\,\% field precision, rising to 98.7\,\% on fields where all four pipelines agree, concentrating manual effort on genuine disagreements. Together, these show that a small, expert-validated annotation effort is enough to begin converting decades of handwritten pottery documentation from Carnuntum into searchable, structured metadata, making archive-scale digitisation feasible for a collection that has so far remained inaccessible to computational analysis.


\section*{Acknowledgements}
This work was carried out within the project \textit{\textbf{\href{https://legion-hsa-2-0.github.io}{LEGION}} (machine \textbf{LE}arnin\textbf{G}-enabled \textbf{I}dentification of archaeological \textbf{O}bjects in the middle da\textbf{N}ube river basin)}. The project is funded by the Austrian Academy of Sciences (OeAW) through the Heritage Science Austria 2.0 programme (grant number \textit{Heritage\_2024-12\_LEGION)}.

\clearpage
\appendix
   \renewcommand{\theHsection}{\Alph{section}}

\section*{Appendix}

The appendix provides supplementary material supporting reproducibility and detailed analysis. \Cref{app:dataset} describes the CENTURIA sample selection strategy, annotation protocol, and data format. \Cref{app:implementation} details the inference setup, model checkpoints, post-processing steps, and LoRA fine-tuning hyperparameters used in all experiments. \Cref{app:evaluation_details} reports full field-level OCR results, the complete adaptation strategy comparison, qualitative OCR examples, and fine-tuned model inference speed. \Cref{app:smart_merge} describes the Model Confidence and Smart Merge procedure with validation results. All prompts are collected in \Cref{app:prompts}. 

\section{Dataset Details}
\label{app:dataset}

\subsection{Sample Selection Strategy}

The sample is drawn from six excavation and field walking campaigns carried out at Carnuntum between 1976 and 2017~\cite{GuglEtAl2024}. Text diversity is captured through handwriting variation across campaigns and documentation periods, character types (alphanumeric, special characters, German terminology, abbreviations such as \textit{Carn.} and \textit{g/red}), and annotation complexity. The dataset comprises eight sampling strata drawn from the following six campaigns:

\paragraph{Excavation campaigns}
\begin{itemize}
    \item \textit{Gstettenbreite 1976}~\cite{GuglEtAl2023,GuglEtAl2021}, \textit{Sector 1--2} -- 43 entries
    \item \textit{Gstettenbreite 1976}~\cite{GuglEtAl2023,GuglEtAl2021}, \textit{Sector 3--4} -- 118 entries
    \item \textit{Gstettenbreite 1976}~\cite{GuglEtAl2023,GuglEtAl2021}, \textit{Sector 5--6} -- 121 entries
    \item \textit{Gstettenbreite 2017}~\cite{GuglEtAl2020} -- 5 entries
    \item \textit{Palastruine 1995}~\cite{GuglEtAl2023} -- 5 entries
    \item \textit{Petroneller Burg (Limesgasse) 2016}~\cite{KonecnyEtAl2021} -- 66 entries
\end{itemize}

\paragraph{Field walking campaigns}
\begin{itemize}
    \item \textit{Gstettenbreite 2017}~\cite{GuglEtAl2019} -- 110 entries
    \item \textit{Käsemacherweide (Gladiatorenschule) 2012}~\cite{GuglRadbauer2017} -- 39 entries
\end{itemize}
Despite sharing a common annotation schema, campaigns differ in recording conventions, particularly regarding find identifiers (\texttt{fn}, \texttt{quadrant}, \texttt{kiste}) and measurement notation (\texttt{RD}/\texttt{BD} vs.\ radius and height), reflecting varied fieldwork practices across projects and time periods.

\subsection{Annotation Protocol}
\label{app:annotation}

\paragraph{Bounding Box Conventions.}
Bounding boxes follow a strict left-to-right, top-to-bottom reading order. Where handwriting is heavily slanted or irregularly spaced, boxes are split into individual words or short phrases. Non-overlapping boxes are preferred over pixel-perfect character enclosure, as region overlap causes character duplication in spatial metrics such as SpACER~\cite{bourne2026charactererrorvector}. In addition to text regions, each scan carries one bounding box for the technical drawing. Where annotations are written across the drawing, text and drawing boxes overlap. This is the only allowed case and is captured as a stratification criterion shown in \Cref{tab:dataset_characteristics}.

\paragraph{Transcription Conventions.}
Transcriptions preserve capitalisation, spacing, special characters, and abbreviations exactly as written. Where spacing is ambiguous, particularly around slashes and parentheses, no space is inserted. Measurements are transcribed with a single space between value and unit, for example, \texttt{12 cm}.

\paragraph{Field Schema.}
Field annotation assigns labels by semantic content rather than spatial position, at the level of semantic units which may span multiple tokens. The seven categories are:

\begin{itemize}
    \item \textbf{Excavation:} contextual find information including provenance (Carn. [\textit{Carnuntum}]), project name, find number (FN [\textit{Fundnummer}]), stratigraphic unit (SE [\textit{Stratigrafische Einheit}]), excavation area (FL [\textit{Fläche}]), quadrant, and crate identifier (\textit{Kiste}).
    \item \textbf{Pottery Form:} vessel typology using standard ceramic vocabulary, e.g., jug (\textit{Krug}), bowl (\textit{Schüssel}), lid (\textit{Deckel}), grinding bowl (\textit{mortarium}).
    \item \textbf{Measurement:} physical dimensions including rim diameter (RD), base diameter (BD), and radius values ($R$, $R_A$, $R_i$, $R_M$), optionally with a percentage indicating preserved rim segment.
    \item \textbf{Artefact Category:} firing technique recorded through standardised abbreviations such as \textit{f/ox GK}, combining firing atmosphere and fabric grade.
    \item \textbf{Publication Type:} typological reference consisting of author name and plate or figure number, e.g., \textit{Gassner 1/10}, \textit{Grünewald 1979 Taf. 33/4}.
    \item \textbf{Surface Treatment:} surface finishing technique, e.g., \textit{Üz}, \textit{Glasur}, \textit{Grießbewurf}.
    \item \textbf{Others:} segments not matching any defined field, including scale annotations ([\textit{Maßstab}] \textit{M~1:1}), recording dates, and draughtsperson initials ([\textit{gezeichnet}] \textit{gez.~MBC}).
\end{itemize}

\subsection{Data Format and Structure}
\label{app:dataformat}

The dataset is provided as two JSON files. The transcription file contains, for each scan, the concatenated ground truth text and one entry per bounding box, storing position $(x, y)$, size $(w, h)$ in pixels, a reading-order index, and a drawing index. The field annotation file stores the same text alongside a structured field dictionary (\texttt{id}, \texttt{text}, \texttt{fields}) following the schema in \Cref{tab:field_examples}.

\section{Implementation Details}
\label{app:implementation}

\subsection{Inference Setup}
\label{app:inference}

All experiments are conducted on NVIDIA GeForce RTX 3090 GPUs (24\,GB VRAM each), as reported in Table~\ref{tab:results}. Models are loaded and executed using the Hugging Face Transformers library (version 5.7.0). All VLMs (olmOCR, olmOCR2, LightOnOCR) use bfloat16 precision, while Florence2 uses float16 and TrOCR runs in float32. Inference is performed with a batch size of~1 per image for all VLMs. TrOCR processes all text regions detected by EasyOCR (CRAFT) within a single image as a variable-size batch. olmOCR and olmOCR2 additionally use Flash~Attention~2 for memory-efficient attention computation.

\subsection{Model Checkpoints}
\label{app:checkpoints}

The following checkpoints are used in all experiments:

\begin{itemize}
    \item \textit{TrOCR}~\cite{trocr_large_handwritten_hf}: \texttt{microsoft/trocr-large-handwritten}
    \item \textit{olmOCR}~\cite{olmocr_hf}: \texttt{allenai/olmOCR-7B-0225-preview}
    \item \textit{olmOCR2}~\cite{olmocr2_hf}: \texttt{allenai/olmOCR-2-7B-1025}
    \item \textit{Florence2}~\cite{florencecommunity_hf}: \texttt{florence-community/Florence-2-large}
    \item \textit{LightOnOCR}~\cite{lightonocr2_hf}: \texttt{lightonai/LightOnOCR-2-1B}
\end{itemize}

\subsection{HTR Post-processing}
\label{app:postprocessing}

Post-processing applies the following cleaning steps to raw model output, in order:

\begin{itemize}
    \item \LaTeX{} markup (e.g. \texttt{\textbackslash frac}, \texttt{\textbackslash text}) is converted to plain text equivalents.
    \item Markdown image references (e.g. \texttt{![...](...)}) are removed.
    \item Hyperlinks and URLs are removed.
    \item Model-generated commentary appended after the transcription is removed.
\end{itemize}
For \textit{olmOCR} specifically, the transcription is extracted from the \texttt{natural\_text} field of the returned JSON object before cleaning. For \textit{TrOCR} and \textit{Florence2}, recognised text segments are concatenated in reading order prior to cleaning.

\subsection{LoRA Fine-tuning}
\label{app:lora}

Both olmOCR2 and LightOnOCR are fine-tuned for 2 epochs on the CENTURIA training set ($n=57$), with results also reported at 1 epoch. LoRA is applied with rank $r=16$ (olmOCR2) and $r=8$ (LightOnOCR), $\alpha=32$, and dropout 0.05. Optimisation targets all linear projection layers using AdamW with learning rate $2\times10^{-4}$, weight decay 0.01, gradient clipping at 1.0, a cosine decay schedule with 10\,\% warmup steps, and batch size 1. Images are resized to a maximum dimension of 1288\,px (olmOCR2) and 768\,px (LightOnOCR).

\subsection{Prompting Strategies}
\label{app:prompting_strategies}

\subsubsection{Zero-shot Instruction Prompt.}
\label{app:prompt:zeroshot}
The model input is augmented with a domain context prompt, which is applied equally to olmOCR2 and LightOnOCR. This replaces the default "Transcribe the document." instruction, providing structured vocabulary lists to guide the disambiguation of handwritten tokens, such as site abbreviations, excavation identifiers (\eg, crate, SE, FL and quadrant numbers), vessel forms, ware types, measurement conventions, surface treatments and publication author abbreviations, as well as other recurring tokens. The full prompt is provided in \cref{app:fig:zeroshot_example}.

\subsubsection{Few-shot Prompt.}
\label{app:prompt:fewshot}
Few-shot examples are image-transcription pairs drawn from the CENTURIA training split. Each example is presented as a multi-turn conversation: The model receives an image alongside the default instruction ("Transcribe the document.") and the expected transcription is provided as the assistant response. For $k \in \{1, 5, 8\}$, the first $k$ examples from an ordered set of 8 are used. Examples are selected via stratified sampling from a pool covering all six campaigns, ensuring representative coverage across diverse excavation contexts, varying field combinations, and different label layouts. The 1-shot example is shown in \cref{app:fig:fewshot_example}.

\subsubsection{Instruction + Few-shot Prompt.}
\label{app:prompt:combined}
The combined strategy uses the domain context prompt (\cref{app:fig:zeroshot_example}) as the task instruction in every conversation turn, including the 8 few-shot demonstration turns described in \cref{app:fig:fewshot_example}.

\subsubsection{LLM Field Extraction Prompt.}
\label{app:prompt:llm}
Field extraction uses Qwen3-8B in no-think mode (\texttt{enable\_thinking=False}), decoding greedily with \texttt{max\_new\_tokens=400}. The system prompt encodes the field schema as: (1) twelve hard rules constraining field assignment and output format, (2) per-field definitions with examples, and (3) a fixed JSON output template. Eight few-shot examples are provided, each providing a corrected transcription along with its expected field annotation. The effect of using corrupted versus corrected few-shot examples is evaluated in \cref{app:effect_fewshot_quality}. The full prompt is provided in \cref{app:fig:field_Extr_prompt}.

\section{Evaluation Details}
\label{app:evaluation_details}
\subsection{Per-field transcription analysis}
\label{app:field_results}

Table~\ref{tab:field_results} reports per-model field-level performance across all five baseline and two fine-tuned models. A notable outlier is olmOCR, which achieves the lowest pottery form EMR (8.6\,\%), omitting the field in 384 of 450 images, while reaching 64.4\,\% measurement ANLS. After LoRA fine-tuning, both models reach near-identical measurement EMR (91.67\,\% each), with olmOCR2~(+) leading on Artefact Category (89.51\,\% vs.\ 85.94\,\%) and Publication Type (86.18\,\% vs.\ 75.34\,\%).

\begin{table}[htb]
\centering
\scriptsize
\setlength{\tabcolsep}{2.5pt}
\caption{\textbf{Field-level OCR performance.} Overall is the
per-document average of GT fields found in raw OCR output (test set,
$n=450$). Measurement reports RD (rim diameter, $n=372$),
evaluated with context-based matching to prevent false positives from
crate and publication numbers. EMR uses exact substring matching;
ANLS uses fuzzy matching ($\tau{\geq}0.5$). $(+)$ indicates LoRA
fine-tuned models. Best result \textbf{bold}, second best
\underline{underlined}.}
\label{tab:field_results}
\begin{tabular}{@{}lcccccccccc@{}}
\toprule
 & \multicolumn{2}{c}{\textbf{Overall}}
 & \multicolumn{2}{c}{\textbf{Pottery}}
 & \multicolumn{2}{c}{\textbf{Measur.}}
 & \multicolumn{2}{c}{\textbf{Art.~Cat.}}
 & \multicolumn{2}{c}{\textbf{Publ.~Type}} \\
\cmidrule(lr){2-3}\cmidrule(lr){4-5}\cmidrule(lr){6-7}
\cmidrule(lr){8-9}\cmidrule(lr){10-11}
\textbf{Model}
 & \textbf{EMR} & \textbf{ANLS} & \textbf{EMR} & \textbf{ANLS}
 & \textbf{EMR} & \textbf{ANLS} & \textbf{EMR} & \textbf{ANLS}
 & \textbf{EMR} & \textbf{ANLS} \\
 & $\uparrow$\% & $\uparrow$\% & $\uparrow$\% & $\uparrow$\%
 & $\uparrow$\% & $\uparrow$\% & $\uparrow$\% & $\uparrow$\%
 & $\uparrow$\% & $\uparrow$\% \\
\midrule
TrOCR
 &  7.13 & 32.81 & 15.78 & 46.86 &  2.42 & 52.00
 &  0.22 & 10.37 &  0.00 & 40.61 \\
olmOCR
 & 13.71 & 30.07 &  8.58 & 17.00 & 21.51 & 64.41
 &  1.79 &  9.51 &  0.27 & 36.26 \\
olmOCR2
 & 30.55 & 57.43 & 37.35 & 68.21 & 79.30 & 92.98
 &  2.46 & 19.31 &  1.90 & 64.07 \\
Florence2
 & 17.39 & 51.13 & 19.49 & 55.07 & 36.83 & 79.12
 &  2.01 & 15.60 &  0.00 & 48.22 \\
LightOnOCR
 & 21.59 & 47.62 & 25.99 & 61.75 & 65.05 & 77.52
 &  1.56 & 15.80 &  0.00 & 46.27 \\
\midrule
olmOCR2~(+)
 & \textbf{92.08} & \textbf{95.29}
 & \textbf{91.18} & \textbf{93.98}
 & \textbf{91.67} & \textbf{96.78}
 & \textbf{89.51} & \underline{95.64}
 & \textbf{86.18} & \textbf{97.08} \\
LightOnOCR~(+)
 & \underline{89.40} & \underline{94.75}
 & \underline{90.95} & \underline{93.91}
 & \textbf{91.67} & \underline{96.92}
 & \underline{85.94} & \textbf{96.23}
 & \underline{75.34} & \underline{95.82} \\
\bottomrule
\end{tabular}
\end{table}

\subsection{Impact of improvement strategies on top-performing models}
\label{app:improvement_strategies_full}

\Cref{tab:improvement_strategies_full} reports the results for all adaptation strategies. Few-shot prompting reduces olmOCR2 error (SpACER-M 5.38\,\% at 5 shots) but collapses LightOnOCR into repetitive output loops (SpACER-M 48.71\,\% at 1 shot, 100\,\% at 5 shots). This behaviour is consistent with known repetition failure modes of autoregressive models~\cite{holtzman2019curious}: the olmOCR authors report repetition as their most common failure and attribute it to inputs that deviate from the format the model was fine-tuned on~\cite{olmocrbench}, and narrow-domain few-shot examples can degrade generation more generally~\cite{tang2025few}. LightOnOCR's RLVR-based training optimises for short, well-formed outputs; multi-image few-shot context is exactly such a deviation, causing degenerate repetition rather than guided transcription. LoRA fine-tuning avoids this instability entirely and dominates all prompting strategies for both models.

\begin{table}[htpb]
\centering
\scriptsize
\caption{\textbf{Impact of adaptation strategies on top-performing
models.} Baseline shows zero-shot performance. Best result
\textbf{bold}, second best \underline{underlined}.}
\label{tab:improvement_strategies_full}
\begin{tabular}{@{}lcccc@{}}
\toprule
\textbf{Model / Strategy}
 & \textbf{SpACER-M} & \textbf{BoW F1}
 & \textbf{EMR}      & \textbf{ANLS} \\
 & $\downarrow$ (\%) & $\uparrow$ (\%)
 & $\uparrow$ (\%)   & $\uparrow$ (\%) \\
\midrule
\multicolumn{5}{l}{\textit{olmOCR2}} \\
\quad Baseline (zero-shot)             & 15.58 & 56.74 & 30.55 & 57.43 \\
\quad Zero-shot instruction prompting  &  6.77 & 83.08 & 74.64 & 84.99 \\
\quad Few-shot (1 shot)                &  8.48 & 72.20 & 51.12 & 79.35 \\
\quad Few-shot (5 shots)               &  5.38 & 84.44 & 68.97 & 87.51 \\
\quad Few-shot (8 shots)               &  5.84 & 85.65 & 75.27 & 87.72 \\
\quad Instruction + few-shot (1 shot)  &  7.29 & 85.52 & 71.67 & 86.39 \\
\quad Instruction + few-shot (5 shots) &  7.64 & 84.39 & 73.06 & 86.10 \\
\quad LoRA fine-tuning (1 epoch)       &  2.33 & 94.57 & 88.72 & 93.83 \\
\quad LoRA fine-tuning (2 epochs)      & \textbf{1.11} & \textbf{96.24}
                                       & \textbf{92.08} & \textbf{95.29} \\
\midrule
\multicolumn{5}{l}{\textit{LightOnOCR}} \\
\quad Baseline (zero-shot)             & 18.25 & 50.98 & 21.59 & 47.62 \\
\quad Zero-shot instruction prompting  & 14.13 & 53.71 & 35.27 & 65.14 \\
\quad Few-shot (1 shot)                & 48.71 & 43.57 & 28.02 & 67.24 \\
\quad Few-shot (5 shots)               & 100.00   & 26.78 & 19.73 & 63.02 \\
\quad Few-shot (8 shots)               & 98.88 & 26.48 & 17.43 & 48.05 \\
\quad LoRA fine-tuning (1 epoch)       &  1.59 & 91.63 & 84.08 & 92.25 \\
\quad LoRA fine-tuning (2 epochs)      & \underline{1.21} & \underline{94.80}
                                       & \underline{89.40} & \underline{94.75} \\
\bottomrule
\end{tabular}
\end{table}

\subsection{Field extraction error analysis}
\label{app:field_extraction}

The error breakdown per field is shown in \cref{fig:extraction_both} of the main paper. \Cref{tab:extraction_errors} details the underlying error patterns and root causes for both pipelines.

\subsubsection{Shared errors.}
The majority of errors are shared by both models (356 of 592, 
60\,\% under REGEX; 494 of 696, 71\,\% under LLM), confirming that failures originate in OCR noise rather than model or extraction method choice. Digit substitutions dominate across fields: crate numbers (Ki\,2/76~$\to$~Ki\,50/76), publication years (Grünewald\,1979~$\to$~1972), and measurement values (rd:\,14~$\to$~rd:\,16) all fail through visually similar handwritten digits. Rare domain tokens (\textit{Eingegl.\,Ker.}, \textit{Üz}) are consistently omitted by both models. Percentage radius measurements (\texttt{r\_a}, \texttt{r\_i}) fail across all four pipelines, as neither model reliably transcribes the \,\% sign.

\subsubsection{Model-specific errors.}
Among errors unique to one model, LightOnOCR is responsible for the majority: 77\,\% under REGEX (182 of 236) and 73\,\% under LLM (147 of 202). Its dominant weakness is Publication Type (67 REGEX, 52 LLM unique errors), where KE inventory references are truncated (Adler-Wölfl\,Taf.\,96\,KE\,2790~$\to$~Taf.\,96) and multi-reference entries are cut off. Under REGEX, it additionally omits Artefact Category and Surface Treatment tokens that fall outside pattern coverage. olmOCR2 produces fewer unique errors but shows a systematic weakness in Pottery Form, truncating compound forms (\textit{Deckel/Schüssel}~$\to$~\textit{Schüssel}) or omitting the field entirely (18 REGEX, 17 LLM unique errors).

\subsubsection{Effect of few-shot example quality.}
\label{app:effect_fewshot_quality}
At inference, the LLM operates on noisy OCR predictions rather than clean transcriptions, so few-shot examples built from corrupted OCR output paired with the correct field annotations match the real inference condition more closely. We therefore compare both variants. \Cref{tab:field_extraction_llm_ablation} compares LLM extraction accuracy using corrupted versus corrected few-shot examples. The largest gain appears on Publication Type, where olmOCR2~(+) improves from 74.53\,\% to 82.11\,\% and LightOnOCR~(+) from 62.87\,\% to 68.83\,\%, suggesting that accurate demonstrations of multi-reference entries help the LLM learn the expected list output format. For all other fields, corrupted and corrected examples show negligible differences. For the fine-tuned models, corrected examples are equal or better on every field; on the zero-shot baselines, the two differ by under one point in either direction. All pipeline results reported in the main paper use corrected few-shot examples.

\begin{table}[t]
\centering
\scriptsize
\setlength{\tabcolsep}{3pt}
\caption{\textbf{Characteristic extraction errors under REGEX and LLM extraction.}
REGEX: 592 errors, 60\,\% shared; LLM: 696 errors, 71\,\% shared.
\textbf{Both}: error shared by olmOCR2~(+) and LightOnOCR~(+).}
\label{tab:extraction_errors}
\resizebox{\linewidth}{!}{%
\begin{tabular}{@{}p{1.5cm} p{1.6cm} p{1.3cm} p{3.2cm} p{2.2cm} p{1.1cm}@{}}
\toprule
\textbf{Field} & \textbf{Error type} & \textbf{Pipeline} 
  & \textbf{GT $\to$ Prediction} & \textbf{Root cause} & \textbf{Scope} \\
\midrule
\multirow{2}{*}{Excavation}
  & Crate digit shift & Both & Ki\,2/76 $\to$ Ki\,50/76       & Digit misread       & \textbf{Both} \\
  & Slash drop        & LLM  & Ki\,1/76 $\to$ Ki\,1776         & Delimiter OCR error & \textbf{Both} \\
\midrule
\multirow{2}{*}{Pottery Form}
  & Misclassif.         & Both & Topf $\to$ Schüssel             & Visual form ambiguity & \textbf{Both} \\
  & Compound truncation & Both & Deckel/Schüssel $\to$ Schüssel  & Prefix dropped        & olmOCR2 \\
\midrule
\multirow{2}{*}{Artefact Cat.}
  & Oxidation confusion & Both & g/red\,GK $\to$ f/ox\,GK       & Firing code OCR error & \textbf{Both} \\
  & Partial extraction  & LLM  & f/ox\,Glasierte\,GK $\to$ f/ox & Token truncation      & olmOCR2 \\
\midrule
\multirow{4}{*}{Publication}
  & Year digit error    & Both  & Grünewald\,\textit{1979} $\to$ \textit{1972} & Single digit OCR    & \textbf{Both} \\
  & Multi-ref confusion & REGEX & Adler-Wölfl\,2010\,\ldots $\to$ Gassner\,3/4 & Ref.\ misassignment & \textbf{Both} \\
  & KE\,no.\ truncation & LLM   & Taf.\,98\,KE\,2847 $\to$ Taf.\,98/           & Ref.\ cut off       & \textbf{Both} \\
  & KE\,ref.\ truncation & REGEX & Adler-Wölfl Taf.\,96\,KE\,2790 $\to$ Taf.\,96 & Ref.\ cut off     & LightOn \\
\midrule
Measurement
  & Value shift         & Both & rd:\,14 $\to$ rd:\,16           & Digit OCR error     & \textbf{Both} \\
\midrule
\multirow{2}{*}{Surface Treat.}
  & Abbrev.\ omission   & Both & Üz $\to$ \textit{null}          & Short token unmatched & \textbf{Both} \\
  & Partial corruption  & LLM  & Üz $\to$ Üz-                    & Trailing char OCR     & LightOn \\
\midrule
Others
  & Noise / omission    & Both & SPA $\to$ (Anm\,|\,SPA) / \textit{null} & Layout debris & \textbf{Both} \\
\bottomrule
\end{tabular}}
\end{table}

\begin{table}[t]
\centering
\footnotesize
\caption{\textbf{Field extraction accuracy (\%) -- LLM extraction with corrupted vs.\ corrected few-shot examples (Qwen3-8B).} Overall is the per-document average across all fields. Excavation aggregates provenance, crate, project, FN, SE, FL, and quadrant (weighted by $n$). $(+)$ denotes LoRA fine-tuned models. \textbf{Bold}: best per column.}
\label{tab:field_extraction_llm_ablation}
\resizebox{\linewidth}{!}{%
\begin{tabular}{@{}lcccccc|cccccc@{}}
\toprule
& \multicolumn{6}{c}{\textbf{LLM -- Corrupted Few-Shots}}
& \multicolumn{6}{c}{\textbf{LLM -- Corrected Few-Shots}} \\
\cmidrule(lr){2-7} \cmidrule(lr){8-13}
\textbf{Model}
  & \textbf{Ovr.} & \textbf{Excav.} & \textbf{Form}
  & \textbf{Ware} & \textbf{Pub.} & \textbf{Surf.}
  & \textbf{Ovr.} & \textbf{Excav.} & \textbf{Form}
  & \textbf{Ware} & \textbf{Pub.} & \textbf{Surf.} \\
\midrule
olmOCR2
  & 29.25 & 29.01 & 34.80 &  1.79 & 0.00 & 0.00
  & 29.12 & 28.93 & 34.34 &  2.23 & 0.27 & 0.00 \\
LightOnOCR
  & 26.79 & 25.49 & 29.00 &  2.01 & 0.27 & 0.00
  & 26.61 & 25.92 & 28.31 &  1.79 & 0.00 & 0.00 \\
\midrule
olmOCR2~(+)
  & 89.71 & 95.79 & 88.63 & 88.39 & 74.53 & \textbf{57.41}
  & \textbf{90.76} & \textbf{95.96} & \textbf{88.86}
  & \textbf{88.84} & \textbf{82.11} & \textbf{57.41} \\
LightOnOCR~(+)
  & 86.68 & 93.99 & \textbf{88.86} & 85.04 & 62.87 & 35.19
  & 87.47 & 94.16 & \textbf{88.86} & 85.04 & 68.83 & 35.19 \\
\bottomrule
\end{tabular}}
\end{table}

\newpage
\subsection{Qualitative transcription examples}
\label{app:field_extraction:qualitative}

\begin{figure}[htbp]
\centering
\begin{subfigure}[b]{0.56\linewidth}
    \includegraphics[width=\linewidth]{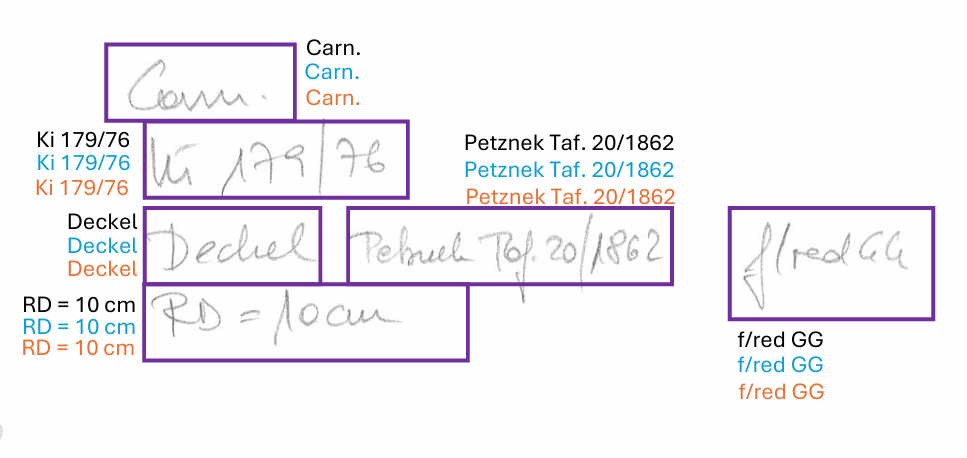}
    \caption{High-quality scan (typical case)}
    \label{fig:qual_high}
\end{subfigure}
\hfill
\begin{subfigure}[b]{0.41\linewidth}
    \includegraphics[width=\linewidth]{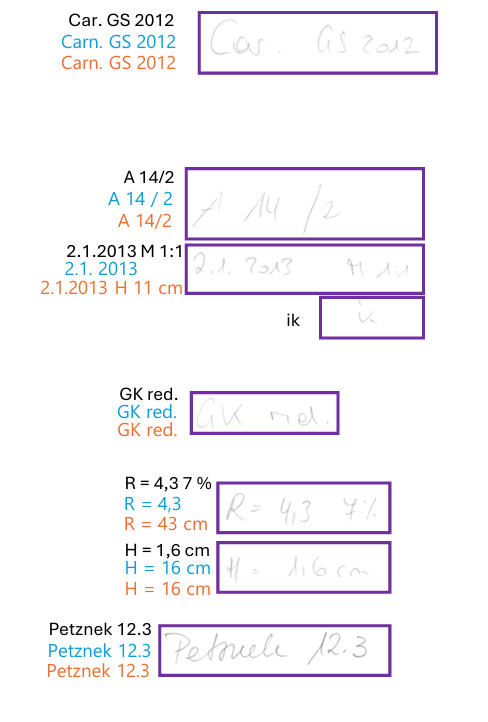}
    \caption{Degraded scan (ink contrast: 43.0, sharpness: 0.5), identified by automated quality assessment (ink contrast, paper noise, Laplacian variance).}
    \label{fig:qual_degraded}
\end{subfigure}
\caption{\textbf{Qualitative OCR examples.} Ground truth (black), olmOCR2~(+) (orange), and LightOnOCR~(+) (blue). On the high-quality scan, both models transcribe all fields correctly. On the degraded scan, low contrast and faded ink cause systematic errors and challenge human annotation as well.}
\label{fig:qual_examples}
\end{figure}

\cref{fig:qual_examples} contrasts a typical high-quality scan with one of the two degraded images identified in the dataset (0.4\,\%). On the high-quality example, both fine-tuned models transcribe all fields correctly, including the \texttt{publication\_type} (\textit{Petznek Taf.\,20/1862}) and \texttt{artefact\_category} (\textit{f/red GG}). On the degraded scan, faded ink and low contrast introduce errors that affect both models and complicate human annotation: \textit{M\,1:1} is misread as \textit{H\,11\,cm} (olmOCR2) and omitted entirely by LightOnOCR, \textit{R\,=\,4{,}3 \,7\,\%} loses the percentage in both models and is additionally read as \textit{R\,=\,43\,cm} by olmOCR2, and \textit{H\,=\,1{,}6\,cm} is read as \textit{H\,=\,16\,cm} by both models. The rarity of such cases in CENTURIA (2 of 507 images) limits their impact on aggregate metrics, but illustrates the difficulty of low-contrast handwritten domain text.

\Cref{tab:qual_fields} shows field extraction results for the two qualitative examples in \cref{fig:qual_examples} across all four model-annotation combinations. On the high-quality scan, all fields are extracted correctly by both models under both methods. On the degraded scan, all four combinations still assign every field to the correct category. However, three fields carry wrong values because the extractors reproduce what the OCR read: \texttt{provenance} is read as \textit{Carn.} although the record shows \textit{Car.}, \texttt{r} loses its percentage in both models and is additionally corrupted to 43 by olmOCR2, and \texttt{h} loses its decimal point in both models (1{,}6\,$\to$\,16). Only \texttt{fn} fails at the extraction stage for LightOnOCR under both methods, as spacing artefacts in the OCR output (\textit{A\,14\,/\,2} vs.\ \textit{A\,14/2}) break the REGEX pattern and cause LLM misassignment. This confirms that OCR-level errors are the primary failure mode, not the annotation method.

\begin{table}[htbp]
\centering
\scriptsize
\setlength{\tabcolsep}{4pt}
\caption{\textbf{Field extraction results for the qualitative examples}
in Fig.~\ref{fig:qual_examples}.
\textbf{R} = REGEX, \textbf{L} = LLM~(Qwen3-8B) extraction;
subscripts denote olmOCR2~(+) and LightOnOCR~(+).
\cmark~correct; \xmark~extracted value shown in italics.}
\label{tab:qual_fields}
\begin{tabular}{@{}p{2.6cm} p{2.1cm} cccc@{}}
\toprule
\textbf{Field} & \textbf{GT}
  & $\mathbf{R}_\mathbf{olm}$ & $\mathbf{L}_\mathbf{olm}$
  & $\mathbf{R}_\mathbf{lit}$ & $\mathbf{L}_\mathbf{lit}$ \\
\midrule
\multicolumn{6}{@{}l}{\textit{High-quality scan — all fields correctly extracted}} \\[1pt]
\midrule
\texttt{provenance}         & Carn.              & \cmark & \cmark & \cmark & \cmark \\
\texttt{kiste}              & Ki\,179/76         & \cmark & \cmark & \cmark & \cmark \\
\texttt{pottery\_form}      & Deckel             & \cmark & \cmark & \cmark & \cmark \\
\texttt{artefact\_category} & f/red\,GG          & \cmark & \cmark & \cmark & \cmark \\
\texttt{publication\_type}  & Petznek Taf.\,20/… & \cmark & \cmark & \cmark & \cmark \\
\texttt{rd}          & 10                 & \cmark & \cmark & \cmark & \cmark \\
\midrule
\multicolumn{6}{@{}l}{\textit{Degraded scan — OCR errors propagate to extracted values}} \\[1pt]
\midrule
\texttt{provenance}         & Car.         & \xmark\,\textit{Carn.} & \xmark\,\textit{Carn.} & \xmark\,\textit{Carn.} & \xmark\,\textit{Carn.}\\
\texttt{project}            & GS\,2012      & \cmark & \cmark & \cmark & \cmark \\
\texttt{fn}                 & A\,14/2       & \cmark & \cmark & \xmark\,\textit{null} & \xmark\,\textit{A\,14\,/\,2} \\
\texttt{artefact\_category} & GK\,red.      & \cmark & \cmark & \cmark & \cmark \\
\texttt{publication\_type}  & Petznek\,12.3 & \cmark & \cmark & \cmark & \cmark \\
\texttt{r}           & 4{,}3 (7\,\%)        & \xmark\,\textit{43} & \xmark\,\textit{43} & \xmark\,\textit{4{,}3} & \xmark\,\textit{4{,}3} \\
\texttt{h}           & 1{,}6         & \xmark\,\textit{16} & \xmark\,\textit{16} & \xmark\,\textit{16} & \xmark\,\textit{16} \\
\bottomrule
\end{tabular}
\end{table}

\newpage
\subsection{Inference speed and output length}
\label{app:inference_speed}

\definecolor{lightsage}{HTML}{F4F2EA}
\definecolor{lightpurple}{HTML}{EEEDFE}
\definecolor{dustypurple}{HTML}{79769E}
\definecolor{darkpurple}{HTML}{5C59A0}

LightOnOCR~(+) is substantially faster than its baseline (1.42\,s vs.\ 2.95\,s per document). This is consistent with an 86\,\% reduction in average output length (72 vs.\ 509 characters, $n=450$). Domain adaptation suppresses verbose artefacts present in baseline outputs (\LaTeX{} notation, Markdown image links, HTML tables, and explanatory commentary). This results in concise, label-like transcriptions. olmOCR2~(+) shows no speed improvement over its baseline (2.26\,s vs.\ 2.19\,s). Its baseline outputs are already concise (77 characters on average), leaving no room for length-based gains after fine-tuning (Table~\ref{tab:output_length}). \cref{app:fig:lightonocr_base_example_output,app:fig:lightonocr_finetune_example_output} illustrate the output reduction for a single document.

\begin{table}[h]
\centering
\caption{\textbf{Average output length per document on the CENTURIA test set ($n=450$).} Values are mean\,$\pm$\,standard deviation across documents. }
\label{tab:output_length}
\begin{tabular}{lr}
\toprule
Model & Avg.\ chars \\
\midrule
olmOCR2         & $77.1 \pm 36.0$   \\
olmOCR2~(+)     & $72.0 \pm 21.4$   \\
\addlinespace
LightOnOCR      & $508.7 \pm 277.9$ \\
LightOnOCR~(+)  & $72.1 \pm 21.8$   \\
\bottomrule
\end{tabular}
\end{table}

\begin{mdframed}[
    linecolor=dustypurple, linewidth=1.2pt,
    backgroundcolor=lightpurple,
    innerleftmargin=8pt, innerrightmargin=8pt,
    innertopmargin=5pt,  innerbottommargin=5pt,
]
\begin{Verbatim}[fontsize=\scriptsize, breaklines]
$\mathcal{C}_{om}$ / $ki\ 1|76$ / $Schiussel\ Possner\ 3/4$ /
$PD = 16cm$ / Flozell / ![image](image_1.png) /
Note: The image contains a simple sketch of a rectangular
object with a vertical slot at one end ...
\end{Verbatim}
\end{mdframed}
\vspace{-0.5em} 
\captionof{figure}{\textbf{LightOnOCR baseline output (2{,}125 chars).}
Verbose output containing \LaTeX{} notation, a Markdown image
link, and model commentary.}
\label{app:fig:lightonocr_base_example_output}

\vspace{30pt}

\begin{mdframed}[
    linecolor=dustypurple, linewidth=1.2pt,
    backgroundcolor=lightpurple,
    innerleftmargin=8pt, innerrightmargin=8pt,
    innertopmargin=5pt,  innerbottommargin=5pt,
]
\begin{Verbatim}[fontsize=\scriptsize, breaklines]
Carn. / Ki 1/76 / Schüssel / Gassner 3/4 / f/ox GK / RD = 16 cm
\end{Verbatim}
\end{mdframed}
\vspace{-0.5em} 
\captionof{figure}{\textbf{LightOnOCR~(+) fine-tuned output (53 chars)}
for the same document as \cref{app:fig:lightonocr_base_example_output}:
concise, label-like transcription.}
\label{app:fig:lightonocr_finetune_example_output}

\newpage
\section{Model Confidence and Smart Merge}
\label{app:smart_merge}
We estimate annotation confidence by cross-checking field predictions across four pipeline combinations (olmOCR2 (+) and LightOnOCR (+), each with REGEX and LLM extraction). Each field is assigned a confidence level based on agreement after string normalisation (\Cref{tab:confidence_levels}). Fields where not all four pipelines agree are resolved by \textit{Smart Merge}: majority vote when 3 of 4 agree (HIGH), field-specific rules for two-against-two splits (INFERRED), or manual review when no rule applies (REVIEW). For two-against-two splits, a field-specific rule selects the more reliable source (\Cref{tab:smart_merge_errors}). These splits can arise either between the two OCR models or between the two extraction methods. For example, if olmOCR2~(+) reads \texttt{rd\,=\,11} while LightOnOCR~(+) reads \texttt{rd\,=\,16} (digit-shift), Smart Merge selects the olmOCR2~(+) value, matching the ground truth. Publication type year errors and percentage radius measurements (\texttt{r\_a}/\texttt{r\_i}) are always sent to review, as no pipeline resolves them reliably. 

A document is auto-accepted once every field is CONFIRMED, HIGH, or INFERRED. This holds for 254 of 450 images (56.4\,\%). Across the full test set, most fields reach full agreement (CONFIRMED: 2{,}524 fields, 98.7\,\% precision), while HIGH and INFERRED together cover only 2.7\,\% of fields (206 of 7{,}650) at lower precision (55.9\,\% and 43.0\,\%). Within the 254 auto-accepted documents, the auto-resolved fields reach 95.5\,\% precision, with the auto-resolved HIGH and INFERRED fields accounting for most of the residual error. The remaining 43.6\,\% of images contain an unresolved conflict and require manual review.

\begin{table}[h]
\centering\scriptsize
\setlength{\tabcolsep}{4pt}
\caption{\textbf{Smart Merge confidence levels.} Precision computed over all 450 test documents. Auto-accepted: 254 images (56.4\,\%); precision 95.5\,\% computed over all fields in auto-accepted documents. NULL: field not present in the record and all four pipelines correctly predict no value (4{,}649 fields).}
\label{tab:confidence_levels}

\resizebox{0.6\linewidth}{!}{%
\begin{tabular}{@{}lrrl@{}}
\toprule
\textbf{Level} & \textbf{Prec.} & \textbf{Fields} & \textbf{Definition} \\
\midrule
CONFIRMED & 98.7\% & 2{,}524 & All 4 pipelines agree \\
HIGH      & 55.9\% &    92   & 3 of 4 pipelines agree \\
INFERRED  & 43.0\% &   114   & 2-vs-2 split, rule resolved \\
\cmidrule{1-4}
REVIEW    &  ---   &   271   & Unresolved conflict \\
\bottomrule
\end{tabular}}
\end{table}

\begin{table}[h]
\centering\scriptsize
\setlength{\tabcolsep}{3pt}
\caption{\textbf{Characteristic pipeline errors and Smart Merge rules.}}
\label{tab:smart_merge_errors}
\resizebox{0.8\linewidth}{!}{%
\begin{tabular}{@{}llll@{}}
\toprule
\textbf{Pipeline} & \textbf{Field} & \textbf{Error type} & \textbf{Resolution} \\
\midrule
REGEX (both)  & Pottery, Art.\ Cat. & Token omission          & Prefer LLM \\
LLM (both)    & Surface Treatment   & Colour hallucination     & Prefer REGEX \\
LLM (both)    & Publication Type    & Multi-ref.\ missed       & Prefer REGEX \\
LightOnOCR    & Excavation, RD      & Higher digit-shift rate  & Prefer olmOCR2 \\
olmOCR2       & Pub.\ Type (year)   & Single digit misread     & Manual review \\
All pipelines & \texttt{r\_a/r\_i}  & \% sign unrecognised     & Manual review \\
\bottomrule
\end{tabular}}
\end{table}

\newpage
\section{Prompts}
\label{app:prompts}

\mdfdefinestyle{promptstyle}{
    linecolor=dustypurple,
    linewidth=1.5pt,
    backgroundcolor=lightpurple,
    innerleftmargin=8pt, innerrightmargin=8pt,
    innertopmargin=6pt,  innerbottommargin=6pt,
}
\mdfdefinestyle{fewshotstyle}{
    linecolor=dustypurple,
    linewidth=1.5pt,
    backgroundcolor=lightpurple,
    innerleftmargin=8pt, innerrightmargin=8pt,
    innertopmargin=6pt,  innerbottommargin=6pt,
}
\mdfdefinestyle{promptstyle_fields}{
    linecolor=dustypurple,
    linewidth=1.5pt,
    backgroundcolor=lightpurple,
    innerleftmargin=8pt, innerrightmargin=8pt,
    innertopmargin=6pt,  innerbottommargin=6pt,
}
\begin{mdframed}[nobreak=true, style=promptstyle]
\begin{Verbatim}[fontsize=\tiny, breaklines]
Transcribe the handwritten text in this image exactly as it appears.
This is an archaeological pottery label from Carnuntum (Roman site in Austria).
Use the following vocabulary to resolve ambiguous handwriting.
Output the text exactly as written. Do not add explanations or structure.

Provenance: 
- Carn.
- Car. 
- CAR 
- Carnuntum

Excavation Identifiers:
- Crate number: Ki X/76
- Find number: FN X
- Stratigraphic Unit: SE X
- Area: FL X
- Survey quadrant: Q XX-YY
- Alphanumeric find number: A-Z X/Y
- Project names: GS, EVN, Survey, Limesgasse

Vessel forms:
Topf, Krug, Kanne, Becher, Faltenbecher, Vorratsgefäß, Nachttopf, Schüssel, Schale, 
Teller, Reibschüssel, Mortarium, Dreifußschale, Räucherschale, Kragenschale, Deckel, 
Sieb, Amphore, Knickwandschüssel

Ware types:
- f/ox GK, g/red GK, f/red GK, g/ox GK, GK Mischbrand
- f/ox GG, f/red GG, g/red GG
- f/red PGW, g/red PGW, f/ox PGW
- f/ox Glasierte GK, Eingegl. Ker. Horreum Ker.
- GK red., GK ox, GK oxid., red. GK, oxid. GK

Measurements (numeric value + cm, optional %):
RD, BD, H, R, R_A, R_i, R_m

Surface treatment:
Üz, Üz a, Üz i, Üz a + i, roter Üz, weißer Üz, Grießbewurf, Glasur, Glasspritzer, Ratterdekor

Publication references (author + year/number/table):
- Common abbreviations: Taf., Abb., KE, Str., Sü, Su
- Gassner
- Petznek
- Grünewald
- Adler-Wölfl
- Horvat
- Stuppner
- Friesinger-Kerchler
- Gattringer-Grünewald

Other recurring tokens:
- SPA
- Fab A.
- Fab B.
- M 1:1
- 1:1
- Gez. MBC 
- ik
- iK 
- Dates: DD.MM.YYYY
\end{Verbatim}
\end{mdframed}
\vspace{-0.5em} 
\captionof{figure}{\textbf{Zero-shot Instruction prompt for OCR.} Domain-specific vocabulary lists guide the model to resolve ambiguous handwriting, covering provenance tokens, excavation identifiers, vessel forms, ware types, measurements, surface treatments, and publication abbreviations.}
\label{app:fig:zeroshot_example}

\begin{mdframed}[style=fewshotstyle]
\begin{Verbatim}[fontsize=\tiny, breaklines]
[
    {
        "image_path": "./dataset/dataset_v2_test_train/train/Excavation Gstettenbreite 1976 Sektor 1-2/005.jpg",
        "transcription": "Carn.\nKi 1/76\nDeckel\nPetznek Taf. 20/1788\nRD = 12 cm\nf/red GK"
    }
]
\end{Verbatim}
\end{mdframed}
\vspace{-0.5em} 
\captionof{figure}{\textbf{1-Shot few-shot example for OCR prompting.} An image-transcription pair from the CENTURIA training set is added to the model input to demonstrate the expected output format and domain conventions.}
\label{app:fig:fewshot_example}

\begin{mdframed}[nobreak=true, style=promptstyle_fields]
\begin{Verbatim}[fontsize=\tiny, breaklines]
SYSTEM_PROMPT = """/no_think

You are an expert in archaeological pottery labels from Carnuntum and OCR error analysis.
Extract structured fields from the prediction text. Copy OCR-corrupted values exactly as they appear.

HARD RULES (never violate these):

1.  If a field is not recognisable in the prediction: null. Do not guess.
2.  NEVER write the string "others" as a field value. It is only valid as a list in the "others" array.
3.  measurement values: extract the NUMBER only and strip all units (cm, mm, μm, am, etc.).
    "14 cm" -> "14"  |  "9,5 cm" -> "9,5"  |  "12mm" -> "12"  |  "22 μm" -> "22"
4.  If a measurement label (H, R, RD, BD, R_A, ...) has NO numeric value -> null.
5.  pottery_form: ONLY from this exact list (OCR variants accepted):
    Topf, Schüssel, Deckel, Krug, Teller, Schale, Becher, Kanne, Sieb,
    Reibschüssel, Räucherschale, Dreifußschale, Dreifuß, Mortarium,
    Faltenbecher, Vorratsgefäß, Nachttopf, Flasche, Knickwandschüssel,
    Käseform, Amphore
    If a token looks like an OCR variant of one of these: assign it.
    If it does not match any: null (NEVER assign a random word).
6.  "Taf." is a plate reference, part of publication_type. NOT pottery_form.
7.  surface_treatment: ONLY Üz / Üz a+i / Üz i / Engobe / Glasur and clear OCR variants.
    Long descriptive phrases (e.g. "mit rotem Überzug Innen und Außen") -> null
    Ware descriptions, material labels, "Mischbrand", "Bemalung" -> null
8.  excavation.kiste: ONLY when the prediction shows a Latin-letter prefix immediately
    followed by a number and /76 (e.g. Ki 131/76, li 54/76, W 60/76, Wi 3/76).
    Greek letters (μ, ω, ρ, γ), math symbols (=, ≠, ∞) are NOT kiste prefixes: null.
    A lowercase Latin letter before a find-style number WITHOUT /76 (e.g. i 18/6): fn, NOT kiste.
9.  excavation.se: pure digits only (e.g. "109", "142"), OR compound strings like "39 - 36 - 09" -> assign as-is.
10. Do NOT include artefact_category token in publication_type.
11. Always output ALL excavation subfields (missing: null).
12. Always output ALL measurement subfields (missing: null).

FIELDS:

- excavation.provenance  : Site token (Carn./Car./Cam., etc.)
- excavation.project     : Campaign name (Survey GB 2017 / GS 2012 / EVN GB 2017 / Limesgasse 2016 ...)
- excavation.fn          : Find number (206, E14/4, 67/95, B 14/5 ...)
- excavation.se          : Stratigraphic unit, pure numbers only
- excavation.fl          : Area/Fläche, pure numbers only
- excavation.quadrant    : Quadrant token (Q 40-12, ...)
- excavation.kiste       : See Hard Rule 8
- pottery_form           : See Hard Rule 5 and 6
- artefact_category      : Ware type (f/ox GK / g/red GK / f/red GG / oxid. GK ...),
                           preserve punctuation exactly as written (e.g. "f/ox GK", not "fox GK").
- publication_type       : Author name + number/plate,
                           "Taf." without a preceding author name -> others,
                           If multiple references -> list of strings.
- surface_treatment      : See Hard Rule 7
- measurement.rd         : Rim diameter, number only (Hard Rule 3)
- measurement.bd         : Base diameter, number only
- measurement.h          : Height, number only
- measurement.r_a        : Outer radius, number only. If followed by N% -> ["value", "% N"]
- measurement.r_i        : Inner radius, number only. If followed by N% -> ["value", "% N"]
- measurement.r_m        : Middle radius, number only
- others                 : Everything not assigned: dates, scale (M 1:1), drafter (Gez MBC),
                           SPA, Fab. A/B, symbols, OCR noise, Greek/math tokens

OUTPUT FORMAT:

Respond with valid JSON only. No prose, no markdown fences, no scratchpad:

{
  "excavation": { "provenance": ..., "project": ..., "fn": ..., "se": ..., "fl": ..., "quadrant": ..., "kiste": ... },
  "pottery_form": ...,
  "artefact_category": ...,
  "publication_type": ...,
  "surface_treatment": ...,
  "measurement": { "rd": ..., "bd": ..., "h": ..., "r_a": ..., "r_i": ..., "r_m": ... },
  "others": [...]
}"""
\end{Verbatim}
\end{mdframed}
\vspace{-0.5em} 
\captionof{figure}{\textbf{System prompt for LLM-based field extraction (Qwen3-8B).} The prompt defines the extraction schema, domain constraints, and output format.}
\label{app:fig:field_Extr_prompt}

\newpage


%
%
\bibliographystyle{splncs04}
\bibliography{main}

@inproceedings{huang2022layoutlmv3,
  author={Yupan Huang and Tengchao Lv and Lei Cui and Yutong Lu and Furu Wei},
  title={{LayoutLMv3: Pre-training for Document AI with Unified Text and Image Masking}},
  booktitle={Proceedings of the 30th ACM International Conference on Multimedia},
  year={2022},
  pages = {4083–4091},
  doi={10.1145/3503161.3548112}
}

@article{transkribus,
author = {Sebastian Colutto and Philip Kahle and G{\"u}nter Hackl and G{\"u}nter M{\"u}hlberger},
title = {{Transkribus. A Platform for Automated Text Recognition and Searching of Historical Documents}},
year = {2019},
pages = {463-466},
doi = {10.1109/eScience.2019.00060},
journal={2019 15th International Conference on eScience (eScience)}}

@misc{puigcerver2018pylaia,
  author = {Joan Puigcerver and Carlos Mocholí},
  title = {{PyLaia}},
  year = {2018},
  publisher = {GitHub},
  journal = {GitHub repository},
  howpublished = {\url{https://github.com/jpuigcerver/PyLaia}},
  commit = {commit SHA},
  note         = {Accessed: 2026-08-13}
}

@INPROCEEDINGS{loravsfullfinetuning,
  author={Hüttner, Lukas and Mayr, Martin and Gorges, Thomas and Wu, Fei and Seuret, Mathias and Maier, Andreas and Christlein, Vincent},
  booktitle={2025 IEEE/CVF Winter Conference on Applications of Computer Vision Workshops (WACVW)}, 
  title={{Low-Rank Adaptation vs. Fine-Tuning for Handwritten Text Recognition}}, 
  year={2025},
  volume={},
  number={},
  pages={1233-1242},
  doi={10.1109/WACVW65960.2025.00146}}

@misc{pypottery_toolkit_gh,
  author       = {Cardarelli, Lorenzo},
  title        = {{PyPottery}: A Collection of Python Tools for Archaeological Ceramics},
  year         = {2025},
  version      = {1.0.2},
  license      = {Apache-2.0},
  url          = {https://github.com/lrncrd/PyPottery},
  note         = {Accessed: 2026-06-26}
}

@misc{mistral3,
  author       = {{Mistral AI}},
  title        = {{Mistral Small 3.1}},
  year         = {2025},
  howpublished = {\url{https://mistral.ai/news/mistral-small-3-1/}},
  note         = {Accessed: 2026-08-13}
}

@article{romein2025assessing,
author = {Romein, C. A. and Rabus, A. and Leifert, G. and Ströbel, Phillip},
year = {2025},
pages = {115-134},
title = {{Assessing advanced handwritten text recognition engines for digitizing historical documents}},
volume = {7},
number={1},
journal = {International Journal of Digital Humanities},
doi = {10.1007/s42803-025-00100-0}
}

@inproceedings{simsa2023docile,
    title={{{DocILE} Benchmark for Document Information Localization and Extraction}},
    author={{\v{S}}imsa, {\v{S}}t{\v{e}}p{\'a}n and {\v{S}}ulc, Milan and U{\v{r}}i{\v{c}}{\'a}{\v{r}}, Michal and Patel, Yash and Hamdi, Ahmed and Koci{\'a}n, Mat{\v{e}}j and Skalick{\`y}, Maty{\'a}{\v{s}} and Matas, Ji{\v{r}}{\'\i} and Doucet, Antoine and Coustaty, Micka{\"e}l and Karatzas, Dimosthenis},
    booktitle = {Document Analysis and Recognition -- ICDAR 2023},
    pages     = {147--166},
    publisher = {Springer Nature Switzerland},
    doi       = {10.1007/978-3-031-41679-8_9},
    year      = {2023}
}

@ARTICLE{garrido2025survey,
  author={Garrido-Munoz, Carlos and Rios-Vila, Antonio and Calvo-Zaragoza, Jorge},
  journal={IEEE Transactions on Pattern Analysis and Machine Intelligence}, 
  title={{Handwritten Text Recognition: A Survey}}, 
  year={2026},
  volume={48},
  number={4},
  pages={4367-4387},
  doi={10.1109/TPAMI.2025.3646002}
  }

@inproceedings{kohut2023finetuning,
      title={{Fine-tuning Is a Surprisingly Effective Domain Adaptation Baseline in Handwriting Recognition}}, 
      booktitle={Document Analysis and Recognition - ICDAR 2023},
      author={Koh{\'u}t, Jan and Hradi{\v{s}}, Michal},
      publisher={Springer Nature Switzerland},
      year={2023},
      pages={269--286},
      isbn={978-3-031-41685-9},
      doi={10.1007/978-3-031-41685-9_17}
}

@inproceedings{kim2022donut,
  title     = {{OCR-Free Document Understanding Transformer}},
  author    = {Kim, Geewook and Hong, Teakgyu and Yim, Moonbin and Nam, JeongYeon and Park, Jinyoung and Yim, Jinyeong and Hwang, Wonseok and Yun, Sangdoo and Han, Dongyoon and Park, Seunghyun},
  booktitle={Computer Vision -- ECCV 2022},
  year      = {2022},
  pages = {498-517},
  isbn={978-3-031-19815-1},
  publisher={Springer Nature Switzerland},
  doi={10.1007/978-3-031-19815-1_29}
}

@inproceedings{wolf2025cm1,
      title={{CM1 -- A Dataset for Evaluating Few-Shot Information Extraction with Large Vision Language Models}}, 
      author={Fabian Wolf and Oliver Tüselmann and Arthur Matei and Lukas Hennies and Christoph Rass and Gernot A. Fink},
    year="2026",
    booktitle="Document Analysis and Recognition --  ICDAR 2025",
    publisher="Springer Nature Switzerland",
    address="Cham",
    pages="23--39",
    doi="10.1007/978-3-032-04617-8_2"
}

@inproceedings{varex2026,
  title   = {{VAREX: A Benchmark for Multi-Modal
             Structured Extraction from Documents}},
  author  = {Barzelay, Udi and Azulai, Ophir and Shapira, Inbar and
             Friedman, Idan and Abo Dahood, Foad and Lee, Madison and
             Daniels, Abraham},
    booktitle = {Proceedings of the IEEE/CVF Conference on Computer Vision and Pattern Recognition (CVPR) Workshops},
    year      = {2026},
    pages     = {7368-7376}
}

@article{Qwen2VL,
  title={{Qwen2-VL: Enhancing Vision-Language Model's Perception of the World at Any Resolution}},
  author={Wang, Peng and Bai, Shuai and Tan, Sinan and Wang, Shijie and Fan, Zhihao and Bai, Jinze and Chen, Keqin and Liu, Xuejing and Wang, Jialin and Ge, Wenbin and Fan, Yang and Dang, Kai and Du, Mengfei and Ren, Xuancheng and Men, Rui and Liu, Dayiheng and Zhou, Chang and Zhou, Jingren and Lin, Junyang},
  journal={arXiv preprint arXiv:2409.12191},
  doi= {10.48550/arXiv.2409.12191},
  year={2024}
}

@article{qwen2.5-VL,
      title={{Qwen2.5-VL Technical Report}}, 
      author={Shuai Bai and Keqin Chen and Xuejing Liu and Jialin Wang and Wenbin Ge and Sibo Song and Kai Dang and Peng Wang and Shijie Wang and Jun Tang and Humen Zhong and Yuanzhi Zhu and Mingkun Yang and Zhaohai Li and Jianqiang Wan and Pengfei Wang and Wei Ding and Zheren Fu and Yiheng Xu and Jiabo Ye and Xi Zhang and Tianbao Xie and Zesen Cheng and Hang Zhang and Zhibo Yang and Haiyang Xu and Junyang Lin},
      year={2025},
      eprint={2502.13923},
      archivePrefix={arXiv},
      primaryClass={cs.CV},
      doi={10.48550/arXiv.2502.13923}, 
    journal={arXiv preprint arXiv:2502.13923},

}

@article{qwen3technicalreport,
      title={Qwen3 Technical Report}, 
      author={An Yang and Anfeng Li and Baosong Yang and Beichen Zhang and Binyuan Hui and Bo Zheng and Bowen Yu and Chang Gao and Chengen Huang and Chenxu Lv and Chujie Zheng and Dayiheng Liu and Fan Zhou and Fei Huang and Feng Hu and Hao Ge and Haoran Wei and Huan Lin and Jialong Tang and Jian Yang and Jianhong Tu and Jianwei Zhang and Jianxin Yang and Jiaxi Yang and Jing Zhou and Jingren Zhou and Junyang Lin and Kai Dang and Keqin Bao and Kexin Yang and Le Yu and Lianghao Deng and Mei Li and Mingfeng Xue and Mingze Li and Pei Zhang and Peng Wang and Qin Zhu and Rui Men and Ruize Gao and Shixuan Liu and Shuang Luo and Tianhao Li and Tianyi Tang and Wenbiao Yin and Xingzhang Ren and Xinyu Wang and Xinyu Zhang and Xuancheng Ren and Yang Fan and Yang Su and Yichang Zhang and Yinger Zhang and Yu Wan and Yuqiong Liu and Zekun Wang and Zeyu Cui and Zhenru Zhang and Zhipeng Zhou and Zihan Qiu},
      year={2025},
      eprint={2505.09388},
      archivePrefix={arXiv},
      primaryClass={cs.CL},
      doi={10.48550/arXiv.2505.09388},
      journal={arXiv preprint arXiv:2505.09388},

}

@misc{olmocrbench,
      title={{olmOCR: Unlocking Trillions of Tokens in PDFs with Vision Language Models}},
      author={Jake Poznanski and Jon Borchardt and Jason Dunkelberger and Regan Huff and Daniel Lin and Aman Rangapur and Christopher Wilhelm and Kyle Lo and Luca Soldaini},
      year={2025},
      eprint={2502.18443},
      archivePrefix={arXiv},
      primaryClass={cs.CL},
      doi={10.48550/arXiv.2502.18443}
}

@misc{olmocr2,
      title={{olmOCR 2: Unit Test Rewards for Document OCR}},
      author={Jake Poznanski and Luca Soldaini and Kyle Lo},
      year={2025},
      eprint={2510.19817},
      archivePrefix={arXiv},
      primaryClass={cs.CV},
      doi={10.48550/arXiv.2510.19817}
}

@inproceedings{xiao2023florence2,
  title={{Florence-2: Advancing a Unified Representation for a Variety of Vision Tasks}},
  author={Xiao, Bin and Wu, Haiping and Xu, Weijian and Dai, Xiyang and Hu, Houdong and Lu, Yumao and Zeng, Michael and Liu, Ce and Yuan, Lu},
    booktitle = {Proceedings of the IEEE/CVF Conference on Computer Vision and Pattern Recognition (CVPR)},
    year      = {2024},
    pages     = {4818-4829},
    doi = {10.1109/CVPR52733.2024.00461}
}

@article{li2021trocr,
      title={{TrOCR: Transformer-Based Optical Character Recognition with Pre-trained Models}},
      journal={Proceedings of the AAAI Conference on Artificial Intelligence},
      volume={37},
      number={11},
      pages={13094--13102},
      author={Minghao Li and Tengchao Lv and Lei Cui and Yijuan Lu and Dinei Florencio and Cha Zhang and Zhoujun Li and Furu Wei},
      year={2023},
      doi={10.1609/aaai.v37i11.26538}
}

@misc{taghadouini2026lightonocr,
      title={{LightOnOCR: A 1B End-to-End Multilingual Vision-Language Model for State-of-the-Art OCR}},
      author={Said Taghadouini and Adrien Cavaillès and Baptiste Aubertin},
      year={2026},
      eprint={2601.14251},
      archivePrefix={arXiv},
      primaryClass={cs.CV},
      doi={10.48550/arXiv.2601.14251}
}

@misc{pypottery,
  title={{PyPotteryScan}},
  author={Lorenzo Cardarelli},
  year={2026},
  howpublished = "\url{https://github.com/lrncrd/PyPotteryScan/}",
  note         = {Accessed: 2026-08-13}
}

@misc{easyocr,
  title={{EasyOCR}},
  author={JaidedAI},
  year={2020},
  howpublished = "\url{https://github.com/JaidedAI/EasyOCR}",
  note         = {Accessed: 2026-08-13}
}

@article{bourne2026charactererrorvector,
      title={{The Character Error Vector: Decomposable errors for page-level OCR evaluation}},
      author={Jonathan Bourne and Mwiza Simbeye and Joseph Nockels},
      year={2026},
      journal={arXiv preprint arXiv:2604.06160},
      doi = {10.48550/arXiv.2604.06160}
}

@article{Bellat2025,
    title = {{Machine Learning Applications in Archaeological Practices: A Review}},
    author = {Bellat, Mathias and Orellana Figueroa, Jordy Didier and Reeves, Jonathan Scott and Taghizadeh-Mehrjardi, Ruhollah and Tennie, Claudio and Scholten, Thomas},
    year = {2025},
    journal = {Journal of Computer Applications in Archaeology},
    volume = {8},
    number = {1},
    pages = {282--321},
    doi = {10.5334/jcaa.201}
}

@article{DiAngelo2022,
    title = {{A Review of Computer-based Methods for Classification and Reconstruction of 3D high-density Scanned Archaeological Pottery}},
    author = {Di Angelo, Luca and Di Stefano, Paolo and Guardiani, Emanuele},
    year = {2022},
    journal = {Journal of Cultural Heritage},
    volume = {56},
    pages = {10--24},
    doi = {10.1016/j.culher.2022.05.001}
}

@article{Gattiglia2025,
    title = {{Managing Artificial Intelligence in Archeology. An overview}},
    author = {Gattiglia, Gabriele},
    year = {2025},
    journal = {Journal of Cultural Heritage},
    volume = {71},
    pages = {225--233},
    doi = {10.1016/j.culher.2024.11.020}
}

@article{Orengo2026,
    title = {{Theory and practice of artificial intelligence in archaeology}},
    author = {Orengo, Hector A. and Berganzo-Besga, Iban and Lumbreras, Felipe},
    year = {2026},
    journal = {Journal of Archaeological Science},
    volume = {190},
    pages = {106571},
    doi = {10.1016/j.jas.2026.106571}
}

@book{Hand2020,
  author    = {Hand, David J.},
  title     = {Dark Data: Why What You Don't Know Matters},
  year      = {2020},
  publisher = {Princeton University Press},
  doi       = {10.1515/9780691198859}
}

@incollection{buechner-matthews_drawing_2022,
	title = {{Drawing and Photography}},
	isbn = {978-2-493207-10-4},
	doi = {10.4000/books.africae.5780},
	booktitle = {Concise {Manual} for {Ceramic} {Studies}},
	publisher = {Africae, Soleb},
	author = {Buechner-Matthews, Saskia and David, Romain},
	editor = {David, Romain},
	year = {2022},
	pages = {71--77}
}

@book{orton_pottery_2013,
	address = {Cambridge},
	edition = {2nd},
	series = {Cambridge {Manuals} in {Archaeology}},
	title = {{Pottery in {Archaeology}}},
	isbn = {978-0-511-92006-6},
	doi = {10.1017/CBO9780511920066},
	publisher = {Cambridge University Press},
	author = {Orton, Clive and Hughes, Michael},
	year = {2013}
}

@inproceedings{9011031,
  author={Biten, Ali Furkan and Tito, Rubèn and Mafla, Andrés and Gomez, Lluis and Rusiñol, Marçal and Jawahar, C.V. and Valveny, Ernest and Karatzas, Dimosthenis},
  booktitle={2019 IEEE/CVF International Conference on Computer Vision (ICCV)},
  title={{Scene Text Visual Question Answering}},
  year={2019},
  pages={4290--4300},
  doi={10.1109/ICCV.2019.00439}
}

@incollection{zvietcovich2015assessment,
  title = {{A 3D Assessment Tool for Precise Recording of Ceramic Fragments Using Image Processing and Computational Geometry Tools}},
  booktitle = {Across Space and Time: Papers from the 41st Conference on Computer Applications and Quantitative Methods in Archaeology},
  author = {Zvietcovich, Fernando and Castaneda, Benjamin and Castillo, Luis Jaime and Saldana, Julio},
  editor = {Traviglia, Arianna},
  year = {2015},
  publisher = {Routledge},
  doi = {10.5117/9789089647153-46}
}

@inproceedings{strobel2020much,
  title={{How Much Data Do You Need? About the Creation of a Ground Truth for Black Letter and the Effectiveness of Neural OCR}},
  author={Str{\"o}bel, Phillip Benjamin and Clematide, Simon and Volk, Martin},
  booktitle={Proceedings of the Twelfth Language Resources and Evaluation Conference},
  pages={3551--3559},
  year={2020},
  doi={10.5167/uzh-197209}
}

@incollection{pendic_use_2024,
	address = {Cham},
	title = {{The {Use} of {3D} {Documentation} for {Investigating} {Archaeological} {Artefacts}}},
	isbn = {978-3-031-53032-6},
	doi = {10.1007/978-3-031-53032-6_2},
	booktitle = {The 3 {Dimensions} of {Digitalised} {Archaeology}: {State}-of-the-{Art}, {Data} {Management} and {Current} {Challenges} in {Archaeological} {3D}-{Documentation}},
	publisher = {Springer International Publishing},
	author = {Pendi{\'c}, Jugoslav and Molloy, Barry},
	editor = {Hostettler, Marco and Buhlke, Anja and Drummer, Clara and Emmenegger, Lea and Reich, Johannes and Stäheli, Corinne},
	year = {2024},
	pages = {9--26}
}

@inproceedings{baek2019characterregionawarenesstext,
  title={{Character Region Awareness for Text Detection}},
  author={Baek, Youngmin and Lee, Bado and Han, Dongyoon and Yun, Sangdoo and Lee, Hwalsuk},
  booktitle={2019 IEEE/CVF Conference on Computer Vision and Pattern Recognition (CVPR)},
  pages={9357--9366},
  year={2019},
  organization={IEEE},
  doi={10.1109/CVPR.2019.00959}}

@article{bergamaschi2022digitalmaktaba,
  author  = {Bergamaschi, Sonia and De Nardis, Stefania and Martoglia, Riccardo and Ruozzi, Federico and Sala, Luca and Vanzini, Matteo and Vigliermo, Riccardo Amerigo},
  title   = {{Novel Perspectives for the Management of Multilingual and Multialphabetic Heritages through Automatic Knowledge Extraction: The DigitalMaktaba Approach}},
  journal = {Sensors},
  volume  = {22},
  number  = {11},
  pages   = {3995},
  year    = {2022},
  doi     = {10.3390/s22113995}
}

@article{mcmanamon2017tdar,
  author  = {McManamon, Francis P. and Kintigh, Keith W. and Ellison, Leigh Anne and Brin, Adam},
  title   = {{tDAR: A Cultural Heritage Archive for Twenty-First-Century Public Outreach, Research, and Resource Management}},
  journal = {Advances in Archaeological Practice},
  volume  = {5},
  number  = {3},
  pages   = {238--249},
  year    = {2017},
  doi     = {10.1017/aap.2017.18}
}

@article{fletcher2023legacydata,
  author  = {Fletcher, Emily C.},
  title   = {{Creating a Software Methodology to Analyze and Preserve Archaeological Legacy Data}},
  journal = {Advances in Archaeological Practice},
  volume  = {11},
  number  = {2},
  pages   = {139--151},
  year    = {2023},
  doi     = {10.1017/aap.2022.44}
}

@article{anichini2020archaide,
  author  = {Anichini, Francesca and Banterle, Francesco and Buxeda i Garrig{\'o}s, Jaume and Callieri, Marco and Dershowitz, Nachum and Dubbini, Nevio and Lucendo Diaz, Diego and Evans, Tim and Gattiglia, Gabriele and Green, Katie and Gualandi, Maria Letizia and Hervas, Miguel Angel and Itkin, Barak and Madrid i Fernandez, Marisol and Miguel Gasc{\'o}n, Eva and Remmy, Michael and Richards, Julian and Scopigno, Roberto and Vila, Lloren{\c c} and Wolf, Lior and Wright, Holly and Zallocco, Massimo},
  title   = {{Developing the ArchAIDE Application: A Digital Workflow for Identifying, Organising and Sharing Archaeological Pottery Using Automated Image Recognition}},
  journal = {Internet Archaeology},
  volume = {52},
  paper = {7},
  year    = {2020},
  doi     = {10.11141/ia.52.7}
}

@article{klein2025autarch,
  author  = {Klein, Kevin and Muller, Antoine and Wohde, Alyssa and Gorelik, Alexander V. and Heyd, Volker and L{\"a}mmel, Ralf and Diekmann, Yoan and Brami, Maxime},
  title   = {{An AI-Assisted Workflow for Object Detection and Data Collection from Archaeological Catalogues}},
  journal = {Journal of Archaeological Science},
  volume  = {179},
  pages   = {106244},
  year    = {2025},
  doi     = {10.1016/j.jas.2025.106244}
}

@article{abele2025archivegpt,
  title={{ArchiveGPT}: A human-centered evaluation of using a vision language model for image cataloguing},
  author={Abele, Line and Anders, Gerrit and Ayd{\i}n, Tolgahan and Buder, J{\"u}rgen and Fischer, Helen and Kimmel, Dominik and Huff, Markus},
  journal={Humanities and Social Sciences Communications},
  volume={13},
  number={1},
  pages={1241},
  year={2026},
  publisher={Palgrave},
  doi={10.1057/s41599-026-08367-6}
}

@article{demjan_laser-aided_2023,
  author       = {Demján, Peter and Pavúk, Peter and Roosevelt, Christopher H.},
  title        = {{Laser-Aided Profile Measurement and Cluster Analysis of Ceramic Shapes}},
  journal      = {Journal of Field Archaeology},
  year         = {2023},
  volume       = {48},
  number       = {1},
  pages        = {1--18},
  doi          = {10.1080/00934690.2022.2128549}
}

@misc{trocr_large_handwritten_hf,
  author       = {Li, Minghao and Lv, Tengchao and Cui, Lei and Lu, Yijuan
                  and Florencio, Dinei and Zhang, Cha and Li, Zhoujun and Wei, Furu},
  title        = {{TrOCR-Large-Handwritten -- Model Checkpoint}},
  year         = {2021},
  howpublished = {\url{https://huggingface.co/microsoft/trocr-large-handwritten}},
  note         = {Fine-tuned on the IAM handwriting dataset, accessed: 2026-08-13},
}

@misc{florencecommunity_hf,
  author       = {{Florence-community}},
  title        = {{Florence-2-Large -- Model Checkpoint}},
  year         = {2024},
  howpublished = {\url{https://huggingface.co/florence-community/Florence-2-large}}
}

@misc{olmocr_hf,
  author       = {{Allen Institute for AI}},
  title        = {{olmOCR-7B-0225-preview -- Model Checkpoint}},
  year         = {2025},
  howpublished = {\url{https://huggingface.co/allenai/olmOCR-7B-0225-preview}},

}

@misc{olmocr2_hf,
  author       = {{Allen Institute for AI}},
  title        = {{olmOCR-2-7B-1025 -- Model Checkpoint}},
  year         = {2025},
  howpublished = {\url{https://huggingface.co/allenai/olmOCR-2-7B-1025}},
}

@misc{lightonocr2_hf,
  author       = {{LightOn AI}},
  title        = {{LightOnOCR-2-1B -- Model Checkpoint}},
  year         = {2026},
  howpublished = {\url{https://huggingface.co/lightonai/LightOnOCR-2-1B}},
  note         = {accessed: 2026-08-13},
}

@article{crosilla2025benchmarking,
title = {Benchmarking large language models for handwritten text recognition},
journal = {Journal of Documentation},
volume = {81},
number = {7},
pages = {334-354},
year = {2025},
issn = {0022-0418},
doi = {10.1108/JD-03-2025-0082},
author = {Giorgia Crosilla and Lukas Klic and Giovanni Colavizza}}

@article{marti2002iam,
  title={{The IAM-database: an English sentence database for offline handwriting recognition}},
  author={Marti, Urs-Viktor and Bunke, Horst},
  journal={International Journal on Document Analysis and Recognition},
  volume={5},
  number={1},
  pages={39--46},
  year={2002},
  doi={10.1007/s100320200071}
}

@inproceedings{sanchez2016read,
  title={{ICFHR2016 competition on handwritten text recognition on the READ dataset}},
  author={S{\'a}nchez, Joan Andreu and Romero, Ver{\'o}nica and Toselli, Alejandro H. and Vidal, Enrique},
  booktitle={2016 15th International Conference on Frontiers in Handwriting Recognition (ICFHR)},
  pages={630--635},
  year={2016},
  doi={10.1109/ICFHR.2016.0120}
}

@inproceedings{huang2019sroie,
  title={{ICDAR2019 competition on scanned receipt OCR and information extraction}},
  author={Huang, Zheng and Chen, Kai and He, Jianhua and Bai, Xiang and Karatzas, Dimosthenis and Lu, Shijian and Jawahar, C.V.},
  booktitle={2019 International Conference on Document Analysis and Recognition (ICDAR)},
  pages={1516--1520},
  year={2019},
  doi={10.1109/ICDAR.2019.00244}
}

@article{rombach2025kie,
  title={{Deep Learning based Key Information Extraction from Business Documents: Systematic Literature Review}},
  author={Rombach, Alexander Michael and Fettke, Peter},
  year={2025},
  journal    = {ACM Computing Surveys},
  volume     = {58},
  number     = {2},
  articleno  = {45},
  pages      = {1--37},
  doi={10.1145/3749369}
}

@inproceedings{
hu2021lora,
title={{Lo{RA}: Low-Rank Adaptation of Large Language Models}},
author={Edward J Hu and Yelong Shen and Phillip Wallis and Zeyuan Allen-Zhu and Yuanzhi Li and Shean Wang and Lu Wang and Weizhu Chen},
booktitle={International Conference on Learning Representations},
year={2022},
url={https://openreview.net/forum?id=nZeVKeeFYf9}
}

@article{romero2013esposalles,
  title={{The ESPOSALLES database: An ancient marriage license corpus for off-line handwriting recognition}},
  author={Romero, Ver{\'o}nica and Forn{\'e}s, Alicia and Serrano, Nicol{\'a}s and S{\'a}nchez, Joan Andreu and Toselli, Alejandro H and Frinken, Volkmar and Vidal, Enrique and Llad{\'o}s, Josep},
  journal={Pattern Recognition},
  volume={46},
  number={6},
  pages={1658--1669},
  year={2013},
  publisher={Elsevier},
  doi={10.1016/j.patcog.2012.11.024}
}

@inproceedings{fornes2017icdar2017,
  title={{ICDAR2017 Competition on Information Extraction in Historical Handwritten Records}},
  author={Forn{\'e}s, Alicia and Romero, Ver{\'o}nica and Bar{\'o}, Arnau and Toledo, Juan Ignacio and S{\'a}nchez, Joan Andreu and Vidal, Enrique and Llad{\'o}s, Josep},
  booktitle={2017 14th IAPR International Conference on Document Analysis and Recognition (ICDAR)},
  volume={1},
  pages={1389--1394},
  year={2017},
  organization={IEEE},
  doi={10.1109/ICDAR.2017.227}
  }

@inproceedings{tarride2023simara,
  title={{SIMARA: A Database for Key-Value Information Extraction from Full-Page Handwritten Documents}},
  author={Tarride, Sol{\`e}ne and Boillet, M{\'e}lodie and Moufflet, Jean-Fran{\c{c}}ois and Kermorvant, Christopher},
  booktitle={International Conference on Document Analysis and Recognition},
  pages={421--437},
  year={2023},
  publisher="Springer Nature Switzerland",
  address="Cham",
  doi="10.1007/978-3-031-41682-8_26"
}

@inproceedings{constum2022recognition,
  title={{Recognition and Information Extraction in Historical Handwritten Tables: Toward Understanding Early 20th Century Paris Census}},
  author={Constum, Thomas and Kempf, Nicolas and Paquet, Thierry and Tranouez, Pierrick and Chatelain, Cl{\'e}ment and Br{\'e}e, Sandra and Merveille, Fran{\c{c}}ois},
  booktitle={International Workshop on Document Analysis Systems},
  pages={143--157},
  year={2022},
  publisher="Springer International Publishing",
  address="Cham",
  doi={10.1007/978-3-031-06555-2_10}
}

@inproceedings{clerice2024catmus,
  title={{CATMuS Medieval: A Multilingual Large-Scale Cross-Century Dataset in Latin Script for Handwritten Text Recognition and Beyond}},
  author={Cl{\'e}rice, Thibault and Pinche, Ariane and Vlachou-Efstathiou, Malamatenia and Chagu{\'e}, Alix and Camps, Jean-Baptiste and Levenson, Matthias Gille and Brisville-Fertin, Olivier and Boschetti, Federico and Fischer, Franz and Gervers, Michael and others},
  booktitle={International Conference on Document Analysis and Recognition},
  pages={174--194},
  year={2024},
  publisher="Springer Nature Switzerland",
  address="Cham",
  doi={10.1007/978-3-031-70543-4_11}
}

@article{cardarelli2025pypotteryink,
  title={{PyPotteryInk: One-step diffusion model for sketch to publication-ready archaeological drawings}},
  author={Cardarelli, Lorenzo},
  journal={Journal of Cultural Heritage},
  volume={74},
  pages={300--310},
  year={2025},
  publisher={Elsevier},
  doi = {10.1016/j.culher.2025.06.016}
}

@article{cardarelli2025pypotterylens,
  title={{PyPotteryLens: An Open-Source Deep Learning Framework for Automated Digitisation of Archaeological Pottery Documentation}},
  author={Cardarelli, Lorenzo},
  journal={Digital Applications in Archaeology and Cultural Heritage},
  pages={e00452},
  year={2025},
  publisher={Elsevier},
  volume = {38},
  doi={10.1016/j.daach.2025.e00452}
}

@article{holtzman2019curious,
  title={{The Curious Case of Neural Text Degeneration}},
  author={Holtzman, Ari and Buys, Jan and Du, Li and Forbes, Maxwell and Choi, Yejin},
  journal={arXiv preprint arXiv:1904.09751},
  year={2019},
  doi={10.48550/arXiv.1904.09751}
}

@article{tang2025few,
  title={{The Few-shot Dilemma: Over-prompting Large Language Models}},
  author={Tang, Yongjian and Tuncel, Doruk and Koerner, Christian and Runkler, Thomas},
  journal={arXiv preprint arXiv:2509.13196},
  year={2025},
  doi={10.48550/arXiv.2509.13196}
}

@misc{wiegmann_zeichenrichtlinien_2018,
	type = {Official guidelines},
	title = {Zeichenrichtlinien des {Landesamtes} für {Denkmalpflege} und {Archäologie} {Sachsen}-{Anhalt} ({LDA})},
	url = {https://www.lda-lsa.de/fileadmin/landesmuseum/alle/pdf/pdf_redaktion/lda_zeichenrichtlinien.pdf},
	language = {de},
	urldate = {2026-06-22},
	publisher = {Landesamt für Denkmalpﬂege und Archäologie Sachsen-Anhalt},
	author = {Wiegmann, Mario},
	year = {2018},
        note = {Accessed: {2026-06-22}}
}

@book{collett_introduction_2017,
	title = {{Introduction to Drawing Archaeological Pottery}},
	isbn = {978-0-948393-25-9},
	language = {en},
	publisher = {Chartered Institute for Archaeologists},
	author = {Collett, Lesley},
    series    = {CIfA Professional Practice Paper },
    volume    = {10},
	year = {2017}
}

@incollection{macginnis_ceramic_2022,
	title = {Ceramic {Illustration}},
	isbn = {978-1-80327-141-5},
	doi = {10.2307/jj.15136034.26},
	urldate = {2026-06-22},
	booktitle = {Laying the {Foundations}: {Manual} of the {British} {Museum} {Iraq} {Scheme} {Archaeological} {Training} {Programme}},
	publisher = {Archaeopress},
	author = {Girotto, Elisa},
	editor = {MacGinnis, John and Rey, Sebastien},
	month = jan,
	year = {2022},
    pages = {213-230},
}

@misc{kosak_instruction_2023,
	type = {Teaching document},
	title = {{Instruction to the {Archaeological} {Drawing} of {Ceramic} {Sherds}. {Summer} {School} in {Umm} {Qais} 2023}},
	url = {https://tutorials.idai.world/},
	urldate = {2026-06-22},
	publisher = {German Archaeological Insttute},
	author = {Kosak, Helga and Borlin, Alessia and Hamel, Hanna and Möller, Heike},
	year = {2023},
    note = {Accessed: 2026-08-13}
}

@misc{okayama_board_education_2018,
  author       = {{Okayama City Buried Cultural Properties Center}},
  title        = {{Shutsudobutsu jissoku manyuaru (entō haniwa-hen)}},
  titleaddon   = {[Manual for measured drawings of excavated objects]},
  year         = {2018},
  type         = {Teaching document},
  publisher    = {Okayama City Board of Education},
  location     = {Okayama},
  langid       = {japanese},
  url          = {https://www.city.okayama.jp/kurashi/cmsfiles/contents/0000005/5424/000334012.pdf},
  urldate      = {2026-06-22},
  note = {Accessed: 2026-06-22}
}

@article{moreno_quixal_2013,
  author       = {Moreno Martín, Andrea and Quixal Santos, David},
  title        = {{Bordes, bases e informes: el dibujo arqueológico de material cerámico y la fotografía digital}},
  journal      = {Arqueoweb: Revista sobre Arqueologia en Internet},
  volume       = {14},
  number       = {1},
  year         = {2013},
  pages        = {178--214},
  langid       = {spanish},
  url          = {https://webs.ucm.es/info/arqueoweb/pdf/14/Moreno178-214.pdf},
  urldate      = {2026-06-22},
  note = {Accessed: 2026-06-22}
}

@incollection{GuglEtAl2024,
  author    = {Gugl, Christian and Wallner, Mario and Pollhammer, Eduard},
  title     = {{Carnuntum -- Eine antike Siedlungsagglomeration an der mittleren Donau}},
  booktitle = {Roman Urban Landscape: Towns and Minor Settlements from Aquileia to the Danube},
  editor    = {Horvat, Jana and Groh, Stefan and Strobel, Karl and Belak, Mateja},
  series    = {Opera Instituti Archaeologici Sloveniae},
  volume    = {47},
  year      = {2024},
  pages     = {377--401},
  doi       = {10.3986/9789610508281_19},
publisher={Zalo{\v{z}}ba ZRC}
}

@article{GuglEtAl2019,
  author  = {Gugl, Christian and Radbauer, Silvia and Wallner, Mario},
  title   = {{Archäologische Prospektion 2012--2017 in der Flur Gstettenbreite -- ein Beitrag zur Entwicklung vorstädtischer Siedlungszonen in Carnuntum}},
  journal = {Carnuntum Jahrbuch},
  volume  = {2018},
  year    = {2019},
  pages   = {47--85},
  doi     = {10.1553/cjb_2018s47}
}

@article{GuglRadbauer2017,
  author  = {Gugl, Christian and Radbauer, Silvia},
  title   = {{Der Oberflächensurvey im Bereich der sog. Gladiatorenschule in Carnuntum. Ein Beitrag zur Siedlungsentwicklung der Südperipherie der Zivilstadt}},
  journal = {Carnuntum Jahrbuch},
  volume  = {2016},
  year    = {2017},
  pages   = {117--148},
  doi     = {10.1553/cjb_2016s117}
}

@article{GuglEtAl2023,
  author  = {Gugl, Christian and Radbauer, Silvia and Wallner, Mario and Pollhammer, Eduard},
  title   = {{Ein Querschnitt durch die Stadt -- Teil 1: Chronologie und Struktur der Carnuntiner Zivilstadt auf Basis von geophysikalischen Messungen und der Notgrabung 1976}},
  journal = {Carnuntum Jahrbuch},
  volume  = {2022},
  year    = {2023},
  pages   = {55--100},
  doi     = {10.1553/cjb_2022s55},
}

@article{GuglEtAl2020,
  author  = {Gugl, Christian and Humer, Franz and Radbauer, Silvia and Schindel, Nikolaus and Wallner, Mario and Zabehlicky, Heinrich},
  title   = {{Archäologische Prospektion und Ausgrabungen in der Flur Gstettenbreite: Gräber und Straßenverläufe im westlichen Vorfeld der Carnuntiner Zivilstadt}},
  journal = {Carnuntum Jahrbuch},
  volume  = {2019},
  year    = {2020},
  pages   = {11--53},
  doi     = {10.1553/cjb_2019s11}
}

@article{GuglEtAl2021,
  author  = {Gugl, Christian and Radbauer, Silvia and Wallner, Mario and Humer, Franz and Pollhammer, Eduard and Neubauer, Wolfgang},
  title   = {{Vor den Toren der Stadt -- Struktur und Entwicklung des westlichen Suburbiums der Carnuntiner Zivilstadt. Neubewertung der Notgrabung 1976 aufgrund der geophysikalischen Messungen 2012--2015}},
  journal = {Carnuntum Jahrbuch},
  volume  = {2020},
  year    = {2021},
  pages   = {37--84},
  doi     = {10.1553/cjb_2020s37},
  urldate = {2024-02-29}
}

@misc{GuglVadeanu2023,
  author    = {Gugl, Christian and Vadeanu, Christopher},
  title     = {Archaeological Research in {Carnuntum}: Selected Data},
  date      = {2023-03-28},
  publisher = {ARCHE},
  url       = {https://hdl.handle.net/21.11115/0000-000F-9C36-5},
  note = {Accessed: 2026-08-01}

}

@article{KonecnyEtAl2021,
  author  = {Konecny, Andreas and Humer, Franz and Radbauer, Silvia and Gugl, Christian and Igl, Roman and Fuchshuber, Nicole},
  title   = {{Zwei Infrastruktureinrichtungen des römischen Carnuntum: der Aquädukt in der Flur Gstettenbreite und die Limesstraße}},
  journal = {Carnuntum Jahrbuch},
  volume  = {2020},
  year    = {2021},
  pages   = {11--36},
  doi     = {10.1553/cjb_2020s11}
}

\end{document}